\documentclass[10pt,twocolumn,letterpaper]{article}

\usepackage{cvpr}              

\usepackage[accsupp]{axessibility}  

\usepackage{pifont}

\usepackage{algorithm}
\usepackage{multirow}
\usepackage{bm}
\usepackage{amsmath}
\usepackage{algpseudocode}
\usepackage{booktabs}

\usepackage{colortbl}
\definecolor{mygray}{gray}{.92}

\newcommand{\nameshort}[1]{Nullu}

\renewcommand{\tilde}{\widetilde}

\def\eqref#1{equation~\ref{#1}}

\def\1{\bm{1}}

\def\vtheta{{\bm{\theta}}}

\def\ve{{\bm{e}}}
\def\vf{{\bm{f}}}

\def\vm{{\bm{m}}}

\def\vo{{\bm{o}}}

\def\vr{{\bm{r}}}

\def\vu{{\bm{u}}}
\def\vv{{\bm{v}}}

\def\vx{{\bm{x}}}
\def\vy{{\bm{y}}}
\def\vz{{\bm{z}}}
\def\vW{{\bm{W}}}

\def\mB{{\bm{B}}}

\def\mE{{\bm{E}}}

\def\mI{{\bm{I}}}

\def\mO{{\bm{O}}}

\def\mU{{\bm{U}}}
\def\mV{{\bm{V}}}
\def\mW{{\bm{W}}}
\def\mX{{\bm{X}}}

\DeclareMathAlphabet{\mathsfit}{\encodingdefault}{\sfdefault}{m}{sl}
\SetMathAlphabet{\mathsfit}{bold}{\encodingdefault}{\sfdefault}{bx}{n}

\newcommand{\E}{\mathbb{E}}
\newcommand{\Ls}{\mathcal{L}}
\newcommand{\R}{\mathbb{R}}

\newcommand{\dpo}{\text{DPO}}
\definecolor{cvprblue}{rgb}{0.21,0.49,0.74}
\usepackage[pagebackref,breaklinks,colorlinks,allcolors=cvprblue]{hyperref}

\def\paperID{7485} 
\def\confName{CVPR}
\def\confYear{2025}

\newcommand\blfootnote[1]{%
  \begingroup
  \renewcommand\thefootnote{}\footnote{#1}%
  \addtocounter{footnote}{-1}%
  \endgroup
}

\title{Nullu: Mitigating Object Hallucinations in Large Vision-Language Models via HalluSpace Projection}

\author{Le Yang$^*$, Ziwei Zheng$^*$, Boxu Chen, Zhengyu Zhao, Chenhao Lin, Chao Shen$^\dag$\\
Xi'an Jiaotong University\\
Xi'an, 710049, China\\
\begin{small}
\begin{tt}  
{ \{yangle15@, ziwei.zheng@stu., chenboxu@stu., zhengyu.zhao@\}xjtu.edu.cn}
\end{tt}
\end{small} \\
{\tt\small \{linchenhao@, chaoshen@mail.\}xjtu.edu.cn}
}

\begin{document}
\maketitle

\blfootnote{$*$ Equal contribution. $\dag$ Corresponding author.}

\begin{abstract}


Recent studies have shown that large vision-language models (LVLMs) often suffer from the issue of object hallucinations (OH). To mitigate this issue, we introduce an efficient method that edits the model weights based on an unsafe subspace, which we call HalluSpace in this paper. With truthful and hallucinated text prompts accompanying the visual content as inputs, the HalluSpace can be identified by extracting the hallucinated embedding features and removing the truthful representations in LVLMs. By orthogonalizing the model weights, input features will be projected into the Null space of the HalluSpace to reduce OH, based on which we name our method Nullu. We reveal that HalluSpaces generally contain prior information in the large language models (LLMs) applied to build LVLMs, which have been shown as essential causes of OH in previous studies. Therefore, null space projection suppresses the LLMs' priors to filter out the hallucinated features, resulting in contextually accurate outputs. Experiments show that our method can effectively mitigate OH across different LVLM families without extra inference costs and also show strong performance in general LVLM benchmarks. Code is released at \url{https://github.com/Ziwei-Zheng/Nullu}.

\end{abstract}    
\section{Introduction}
\label{sec:intro}

Recent advancements in large language models (LLMs)~\cite{openai2023gpt4,touvron2023llama,touvron2024llama3} have spurred rapid developments in large vision-language models (LVLMs), such as GPT-4V~\cite{openai2023gpt4}, Gemini~\cite{team2023gemini}, LLaVA~\cite{liu2024visual} and mPLUG-Owl2~\cite{ye2024mplug}. By integrating a vision encoder and fine-tuning on multimodal instruction-following datasets, LVLMs exhibit a strong ability to interpret and convert complex visual patterns into coherent linguistic representations. Whereas, the widespread use of these LVLMs has shed light on their limitations---object hallucinations (OH).


\begin{figure}[t]
    \centering
    \includegraphics[width=\linewidth]{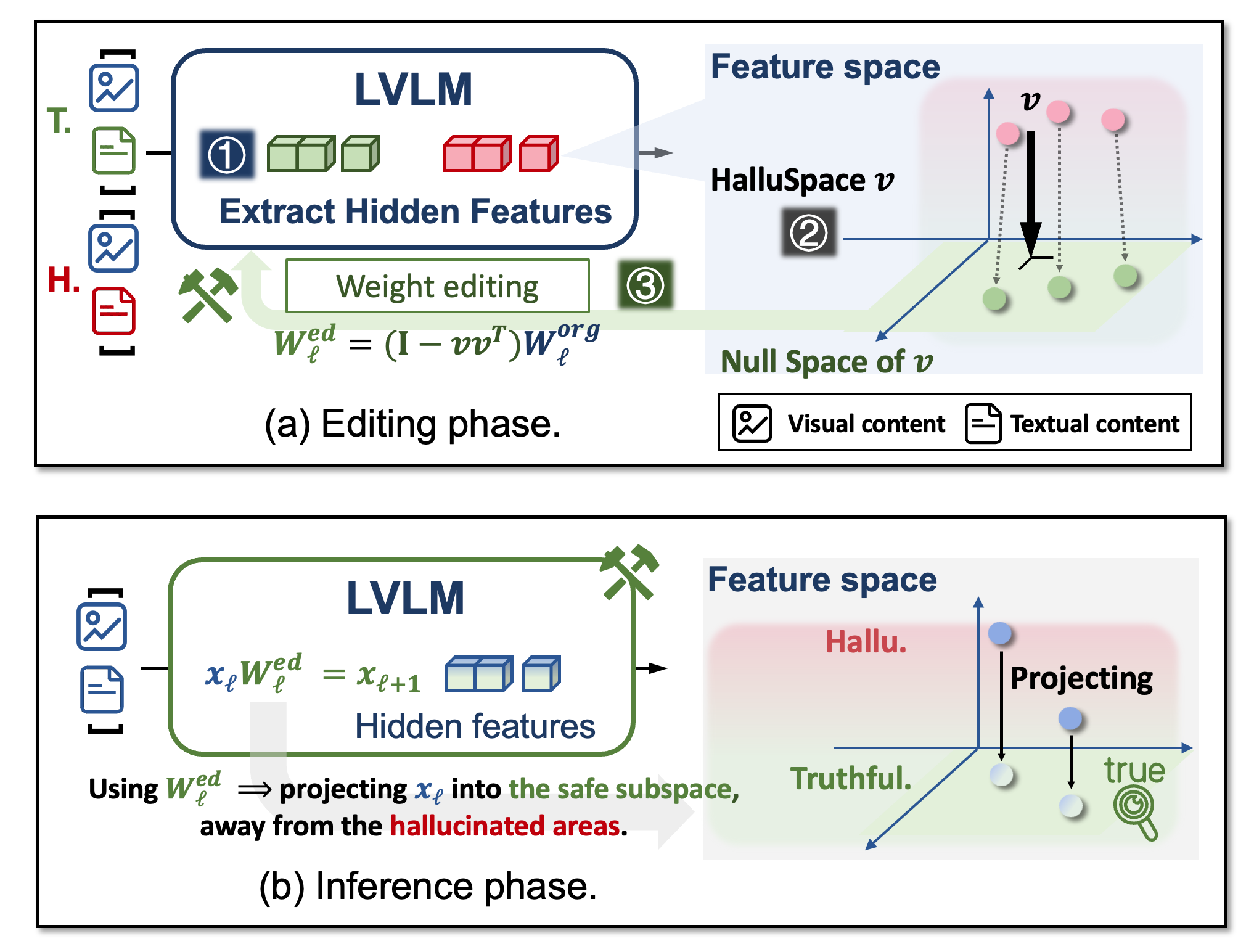} 
    \vspace{-20pt}
    \caption{An illustration of \nameshort{}. (a) In the editing phase, \nameshort{} will \ding{172} Extract hidden features of truthful (\textcolor[rgb]{0.21,0.54,0.25}{\textbf{T.}}) and hallucinated (\textcolor[rgb]{0.75,0.13,0.09}{\textbf{H.}}) inputs. \ding{173} Explore a low-rank HalluSpace $\vv$ in the feature space by contrasting the differences between \textcolor[rgb]{0.21,0.54,0.25}{\textbf{T.}} and \textcolor[rgb]{0.75,0.13,0.09}{\textbf{H.}} features. \ding{174} Edit the model weights by projecting them to the null space of $\vv$. (b) In the inference phase, using the edited weights equals to project input features into the safe subspace, away from the hallucinated areas, leading to contextually accurate outputs.}
    \label{fig1}
    \vspace{-15pt}
\end{figure}

Object Hallucination (OH) refers to the phenomenon where LVLMs generate text that inaccurately represents the actual objects in the accompanying image~\cite{liu2023mitigating,bai2024hallucination,jiang2024hallucination,leng2024mitigating,li2023evaluating}. Considering the wide application of LVLMs in different fields, hallucinations in generated content can lead to misinformation or even harmful decisions. Therefore, mitigating OH is crucial for AI safety and trustworthiness. Recent works propose to address OH by either performing fine-tuning in an end-to-end manner~\cite{liu2023mitigating,jiang2024hallucination,kim2023exposing} or using post-processing methods to modify model outputs~\cite{leng2024mitigating,zhang2024debiasing,zhou2023analyzing,chen2024halc}, which show effectiveness in mitigating OH for modern open-source LVLMs.


Despite significant efforts, the underlying causes of OH within the models remain unexplored. This study investigates this issue by systematically examining the intrinsic distinctions between truthful and hallucinated responses, focusing on feature spaces. Our empirical results on open-source LVLMs reveal a low-rank subspace in the model parameter space, which we term ``HalluSpace'', representing the distinction between truthful and untruthful feature distributions, and can be a factor triggering OH.



Building on this finding, we mitigate OH by projecting the features away from the detected HalluSpaces. Namely, we project the original model weights to the null space of HalluSpaces, based on which we build a novel method called \nameshort{} (shown in Figure~\ref{fig1}). Our method first applies Principal Component Analysis (PCA) to extract the low-dimensional subspaces (HalluSpaces) from the difference matrix between truthful and hallucinated features. Then, by orthogonalizing model weights with respect to the HalluSpaces, \nameshort{} effectively nullifies OH representations for inputs, filtering out these untruthful features and resulting in contextually more accurate outputs.


We also found that these HalluSpace are closely tied to the priors in LLMs, which is primary causative factors of OH in~\cite{leng2024mitigating}. Therefore, projecting model weights to the null space of HalluSpace offers a distinct but straightforward approach to reducing these priors. Furthermore, we establish connections between \nameshort{} and Direct Preference Optimization (DPO)~\cite{rafailov2024direct}, showing that both of these two methods attempt to achieve a similar goal to avoid OH.

Through comprehensive experiments, we demonstrate the effectiveness of \nameshort{} in mitigating OH without compromising the models’ general performance. Notably, \nameshort{} needs no additional fine-tuning procedure and, introduces no extra inference costs compared to most post-hoc based methods~\cite{chen2024halc,leng2024mitigating,zhang2024debiasing}, making \nameshort{} much more efficient in alleviating OH. Our main contributions are as follows:
\begin{enumerate}
    \item We propose a novel method named \nameshort{} for OH mitigation. By extracting HalluSpaces and orthogonalizing the model weights, \nameshort{} can effectively mitigate OH with no extra inference cost. Our method demonstrates the feasibility and effectiveness of mitigating OH by editing weight based on representation learning. 
    \item We show that HalluSpaces appear to form a prior closely related to the priors in LLMs causing OH. Therefore, our method coincides with previous studies debiasing LVLMs to reduce OH at outputs. Moreover, we show the connection between \nameshort{} and DPO, which can provide nuanced insights, broadening the understanding of potential causes behind unsafe behaviors within LVLMs.

    \item Experiments show that using \nameshort{} achieves consistent improvements with zero-extra costs on the evaluated tasks across different open-source LVLMs, demonstrating the effectiveness and efficiency of \nameshort{}. 
\end{enumerate}

\section{Related Work}
\label{sec:related}

\subsection{Large Visual-Language Models (LVLMs)}

Based on the development of LLMs, such as LLaMA~\cite{touvron2023llama2} and Vicuna~\cite{vicuna2023}, large vision-language models (LVLMs) have made significant advancements in recent years. Early works, such as BLIP~\cite{li2022blip,li2023blip} and BERT-based VLMs~\cite{devlin2018bert,liu2019roberta}, have successfully adapted LLMs to visual tasks, demonstrating notable generative capabilities and in-context learning abilities. These advancements typically involve connecting a vision encoder to an LLM through various fused modules, such as linear projection~\cite{liu2024visual} or a Q-former~\cite{zhu2023minigpt4}. With the adoption of visual instruction tuning techniques, LVLMs~\cite{liu2024visual,zhao2024stitch} have further enhanced their abilities, enabling language generation models to perform complex image understanding and reasoning tasks, resulting in a series of modern well-performed LVLMs, such as LLaVA~\cite{liu2024visual,liu2023improved}, Mini-GPT4~\cite{zhu2023minigpt4}, mPLUG-Owl~\cite{ye2024mplug,ye2024mplugowl3longimagesequenceunderstanding}, Shikra~\cite{chen2023shikra}, InternVL~\cite{chen2024internvl} and Qwen-VL~\cite{bai2023qwenllm,bai2023qwen}. However, despite these successes, recent LLMs and LVLMs still face security issues~\cite{wang2024model,zhao2024diver,jing2024fgaif}, such as hallucination~\cite{bai2024hallucination}, 


\subsection{Mitigation of Object Hallucination}

Object hallucination occurs when the model generates semantically coherent texts misaligned with actual objects in the accompanying image~\cite{rohrbach2018object,pmlr-v235-wu24l,wu2024logical,huang2024visual}. Various approaches have been proposed to address this issue. Considering a possible cause of hallucination lies in data biases and the knowledge gap between visual and linguistic information, recent studies propose to fine-tune the LVLMs for robustness~\cite{liu2023mitigating,gunjal2024detecting}, cross-modality matching~\cite{jiang2024hallucination,kim2023exposing} or preference alignments~\cite{sun2023aligning,chen2024dress}. While these methods are effective, they are notorious for the demands of substantial computational resources for training and fine-tuning.

\begin{figure*}[t!]
    \centering
    \includegraphics[width=\linewidth]{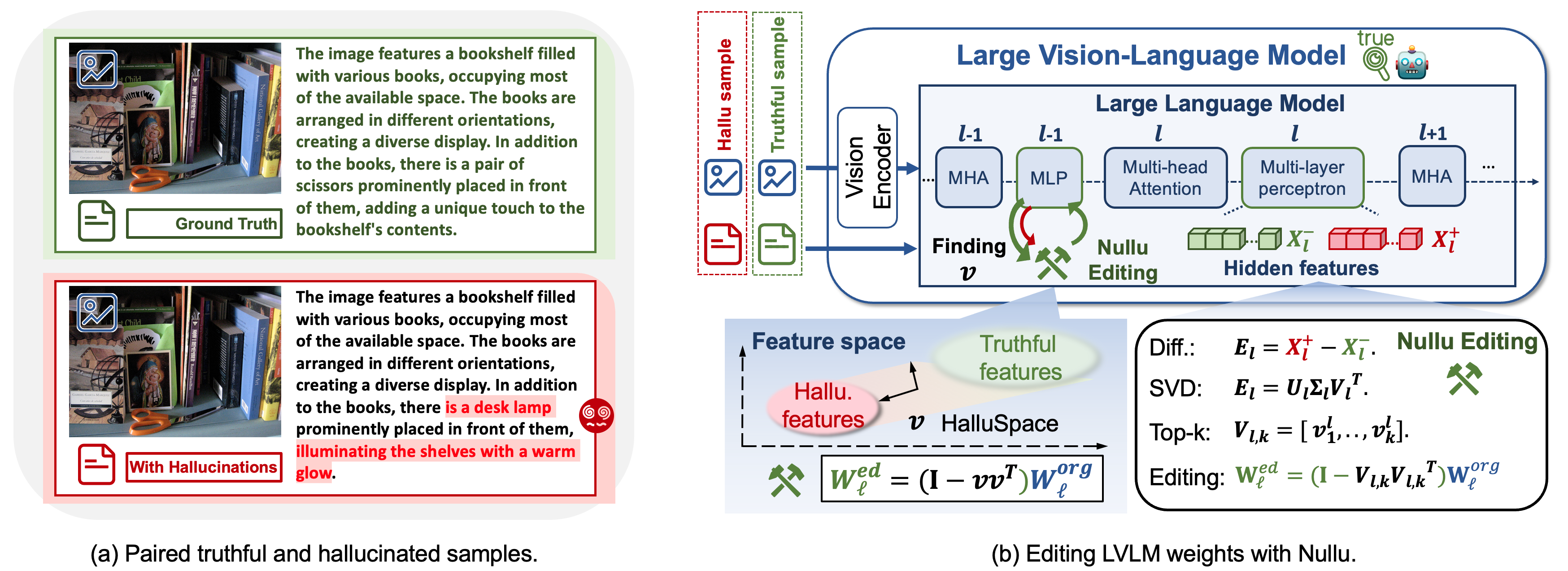}
    \vspace{-15pt}
    \caption{An overview of \nameshort{}, which identifies the HalluSpaces to edit model weights for LVLMs. (a) The paired truthful and hallucinated samples. (b) \nameshort{} first calculates the difference matrix of hidden features for the paired samples and then conducts the SVD to find the main directions of the difference as the HalluSpace. Then \nameshort{} projects the original MLP's weights to the null space of the HalluSpace. This procedure will be repeated for a series of layers, $\{\ell\}$, in the LLM of an LVLM.
    }
    \label{fig:all}
    \vspace{-10pt}
\end{figure*}

Researchers also propose post-processing methods to avoid high computational costs, which apply new strategies or external tools to modify the response. LURE~\cite{zhou2023analyzing} trains a reviser to edit the possible hallucinated words in the responses. Other methods incorporate an external visual model improved answers~\cite{yin2023woodpecker,biten2022let,chen2024halc}. Moreover, recent studies have revealed that the LVLMs are often affected by the strong LLM's priors and visual uncertainty, leading to less attention to the visual contexts and resulting in OH~\cite{leng2024mitigating,liu2024paying,zhang2024debiasing,zhu2024ibd,huang2024opera,favero2024multi}. 

Compared to these works, \nameshort{} mitigates OH by a new editing strategy that contrasts the truthful and hallucinated features, resulting in an effective and efficient solution for OH issues.


\subsection{Feature steering}
In the LLM and LVLM fields, some works use a similar approach of measuring the ``contrast" between the internal features of positive and negative samples and then modifying features at inference time using this ``contrast" to adjust the model's behavior. E.g., ITI~\cite{li2024inference}, ActADD~\cite{turner2023activation}, and VTI~\cite{liureducing} shift the features via an obtained direction for truthfulness, detoxification, and OH mitigation, respectively. Our method projects the features for editing. Although recent work~\cite{uppaal2024detox} also proposes the null space projection strategy for editing, its target is the detoxification of LLMs, which differs from ours.

\section{Method}
\label{sec:method}

The framework of \nameshort{} is illustrated in Figure~\ref{fig:all}. We first describe how to construct the data pairs of truthful and hallucinated text prompts, then introduce how \nameshort{} extracts the HalluSpace to edit the model weights. We further provide an in-depth analysis by decoding the HalluSpaces, and discuss the connection between \nameshort{} and DPO. 



\subsection{Data-pair construction} 


We begin with constructing paired vision-language inputs as shown in Figure~\ref{fig:all} (a). The two inputs have the same image but different text prompts, where one contains the truthful ground truth, $x_i^{-}$, that accurately describes the objects in the image, and the other contains hallucinated descriptions, $x_i^{+}$. The whole dataset can be represented by $\mathcal{D} = {(x_i^{+}, x_i^{-})}_{i=1}^N$. 

As stated in~\cite{zhou2023analyzing}, the hallucinatory descriptions can be generated by modifying the accurate descriptions using GPT-3.5. These adjustments are guided by factors related to object hallucination, including co-occurrence, object uncertainty, and object position. Then, we can obtain the data pair whose sample contains an image with a GT description and the same image with a hallucinated description by substituting the objects in the GT description with the most likely hallucinated ones.

\subsection{\nameshort{}} 




\paragraph{Exploring HalluSpaces.}

At each model layer $\ell$ in the LLM of the LVLM, where $\ell \in {L_0 \dots L}$, we compute embedding features, denoted as $\vx_{i,\ell}^{+}$ and $\vx_{i,\ell}^{-}$ for the hallucinated and truthful features, respectively. We then average each sample over the token dimension and stack these embedding features into matrices $\mX_\ell^{+}$, $\mX_\ell^{-} \in \R^{N \times D}$, where $D$ is the embedding dimension. Then the difference matrix $\mE_\ell$ at $\ell$-th layer can be calculated by:
\begin{align}
    \mE_\ell =  \mX_\ell^{+} - \mX_\ell^{-}\; , \; \; \mE_\ell \in  \mathbb{R}^{N\times D}.
\end{align}
We conduct the principal component analysis for $\mE_\ell $ via singular value decomposition (SVD). As shown in Figure~\ref{fig:all} (b), this matrix contains the underlying differences between the truthful and hallucinated embeddings in the feature space. Therefore, by exploring the structure of $\mE_\ell$, we can find the main directions as a low-rank approximation of $\mE_\ell$, based on which we can effectively eliminate the hallucinated features. Formally, the SVD can be conducted by 
\begin{align}
    \mE_\ell = \mU_\ell \mathbf{\Sigma}_\ell \mV_\ell^\top, \; \mU_\ell\in \mathbb{R}^{N\times N}, \; \mV_\ell\in \mathbb{R}^{D\times D}, 
\end{align}
where $\mathbf{\Sigma}_\ell$ is the diagonal matrix containing the singular values with the descending sort. 

Next, we pick the right singular vectors with top-$k$ singular values, $\vv_1^\ell, \vv_2^\ell, \ldots, \vv_k^\ell$, which are the first $k$ column vectors of $\mV_\ell$. These directions represent the main difference between the truthful and hallucinated features and, therefore, can be considered the directions in model weight spaces that trigger the hallucinated descriptions. Therefore $\mV_{\ell,k} = [\vv_1,..., \vv_k] \in \mathbb{R}^{D\times k}$ is the target HalluSpace.

\begin{figure*}[t!]
    \centering
    \includegraphics[width=\linewidth]{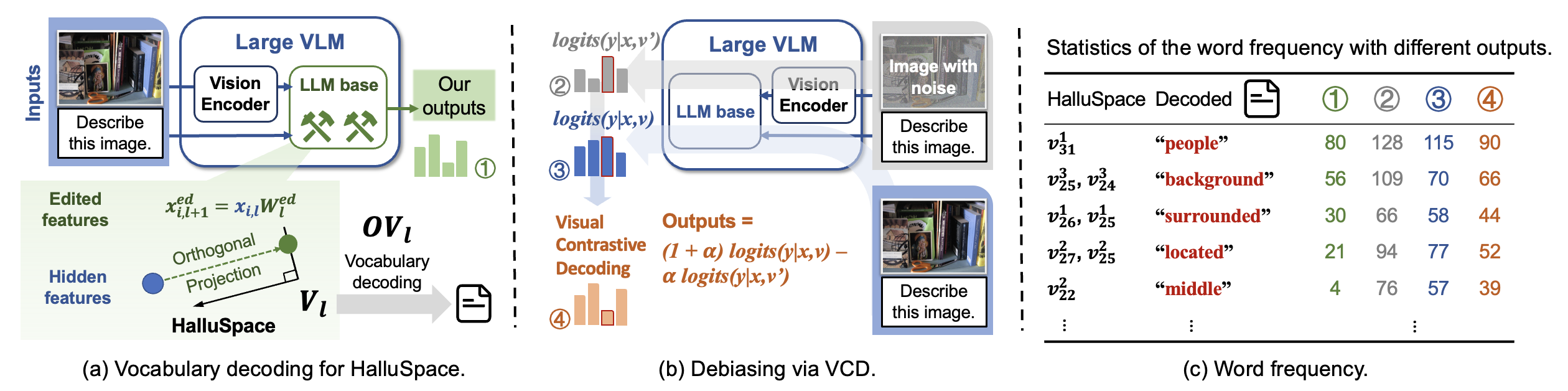}
     \vspace{-19pt}
    \caption{The relation between \nameshort{} and other debiasing methods~\cite{leng2024mitigating,zhang2024debiasing}. The inference procedure of (a) ours and (b) VCD~\cite{leng2024mitigating}. (c) The statistics of the word frequency in outputs of \nameshort{} (\textcolor[rgb]{0.21,0.54,0.25}{\ding{172}}), with LLM priors (\ding{173}), LLaVA (\textcolor[rgb]{0.2,0.31,0.66}{\ding{174}}) and VCD (\textcolor[rgb]{0.7,0.37,0.16}{\ding{175}}).    }
    \label{fig:debias}
    \vspace{-10pt}
\end{figure*}
Usually, with a $k \ll D$, the HalluSpace can be well defined, which means the hallucinated descriptions are possible within a low-rank dimensionality. We then show how to edit the model weights to mitigate OH in LVLMs.

\vspace{-5pt}
\paragraph{Editing the model weights.} As the $\mV_{\ell,k}$ represents the main different directions between the truthful and untruthful data distributions, we can effectively remove the hallucination information from the model features by projecting them to the null space of HalluSpace, which can be achieved by orthogonalizing the model weights with respect to $\mV_{\ell,k}$. The null space of $\mV_{\ell,k}$ can be presented by $(\mI - \mV_{\ell,k} \mV_{\ell,k}^\top)$\footnote{The proof can be found in supplementary materials.}.

Therefore, we project the original weights of MLP at $\ell$-th layer $\mW_{\ell}^\text{org}$ to the null-space of $\mV_{\ell,k}$ with 
\begin{align} 
\label{eq:edit}
    \mW_{\ell}^\text{ed} =  (\mI - \mV_{\ell,k} \mV_{\ell,k}^\top) \; \mW_{\ell}^\text{org}
     \; .
\end{align}
With the $\mW_{\ell}^\text{ed}$, \nameshort{} can effectively mitigate OH while generating responses. As the new weights can be reloaded into the model, there is no extra cost during inference, making our method computationally efficient.

\begin{algorithm}[htp]
   \caption{\nameshort{}}
   \label{alg}
\begin{algorithmic}
   \State \textbf{Input:} Paired data $\mathcal{D}$, LVLM $\mathcal{M}$, target layers $\{\ell\}$, rank number $k$.
   \State \textbf{Output:} Edited LVLM $\mathcal{M}^{ed}$.
   \For {$\ell$ in $\{\ell\}$}
   \State Calculating $\mX_\ell^{+}$ and $\mX_\ell^{-}$ \Comment{Hidden features} 
   \State $\mE_\ell =  \mX_\ell^{+} - \mX_\ell^{-}$
   \State $\mE_\ell = \mU_\ell \mathbf{\Sigma}_\ell \mV_\ell^\top$ 
   \State $\mV_{\ell,k} = [\vv_1,..., \vv_k]$ \Comment{Selecting Top-$k$} 
   \State $\mW_{\ell}^\text{ed} =  (\mI - \mV_{\ell,k} \mV_{\ell,k}^\top) \; \mW_{\ell}^\text{org}$ \Comment{Editing $\ell$-th MLP} 
   \EndFor
\end{algorithmic}
\end{algorithm}
\vspace{-10pt}
\paragraph{\nameshort{}.} We summarize \nameshort{} in Algorithm~\ref{alg}. Our method contains two key parameters: the indices of layers needed to be editing $\{\ell\}$ and the selected top-$k$ singular vectors. Although we do not have a theoretical argument for the best values, we explore their effects experimentally and determine optimal values via a standard hyper-parameter sweep.

\subsection{Decoding information in HalluSpace}

We provide a case study to explore how the proposed \nameshort{} mitigates OH. We conduct the analysis experiments on CHAIR~\cite{rohrbach2018object} with LURE~\cite{zhou2023analyzing}, which provides the ground truth and possible hallucinated responses with each visual input. The introduction of the dataset is provided in Section~\ref{sec:exp}. We use LLaVA-1.5-7B~\cite{liu2023improved} for our evaluation.

\vspace{-5pt}
\paragraph{Mapping the subspace back to vocabulary.} 
Following previous works~\cite{geva2021transformer,uppaal2024detox}, to explore the information behind the learned vectors $\vv_i \in \R^D$, a common approach is to decode the embeddings with $\mO  \vv \in \R^{|\mathcal{V}|}$, where $\mO = [\vo_1, \ldots, \vo_{|\mathcal{V}|}]^\top \in \R^{|\mathcal{V}| \times D}$ and $\mathcal{V}$ denotes the vocabulary. We then sort $\mO \vv$ in ascending order, find the top-$m$ indices, and use the corresponding words to interpret $\vv$. The following analysis is performed at each layer $\ell$, while we will drop $\ell$ to avoid the notational burden.

Using \nameshort{}, we can extract the $\mV$ at different layers and then decode it via $\mO \vv$ to explore the internal information behind $\mV$. For better interpretation, we only select the most representative results\footnote{The complete decoding results of $\mV$ and the reason why we select the most representative results are provided in Supplementary materials.}, containing high-frequent words like ``background'' and ``middle'', shown in Supplementary materials. As far as we know, projecting the model weights orthogonally to $\mV$ can reduce the semantic information of the decoded words during responding. However, why does such a procedure mitigate OH?

\vspace{-8pt}
\paragraph{Debiasing LVLM from LLM prior.} 
Recent studies~\cite{leng2024mitigating} have revealed that hallucinated responses of LVLMs can be attributed to the strong LLMs' priors. Therefore, following the settings in~\cite{leng2024mitigating}, we collect the outputs of LLaVA-1.5 with normal inputs (\textcolor[rgb]{0.2,0.31,0.66}{\ding{174}}) and visually distorted inputs (\ding{173}), as shown in Figure~\ref{fig:debias} (b). The statistical results of word frequency of \ding{173} and \textcolor[rgb]{0.2,0.31,0.66}{\ding{174}} show that LLMs indeed bias on some words. And techniques proposed to calibrate the outputs by reducing these detected LLM priors~\cite{leng2024mitigating} (resulting in the results \textcolor[rgb]{0.7,0.37,0.16}{\ding{175}}), or, enhancing the visual information~\cite{zhu2024ibd,liu2024paying,huang2024opera} during inference to remove these priors in LLMs.


\vspace{-8pt}
\paragraph{Interpreting HalluSpaces.} From the results of \ding{173} in Figure~\ref{fig:debias} (b) and the results in Table 1 in Supplementary materials, we see that the decoded words from HalluSpace overlap with the high frequent words when with the distorted visual inputs (\ding{173}), indicating that HalluSpace $\mV$ might contain similar latent information as LLM's prior. Moreover, by visual contrastive decoding, we see that these LLM's ``preferred'' words are reduced (see \textcolor[rgb]{0.7,0.37,0.16}{\ding{175}}), which has a similar trend to \nameshort{} (\textcolor[rgb]{0.21,0.54,0.25}{\ding{172}}). Therefore, we indicate that \nameshort{} projects the features to the null space of the HalluSpace $\mV$, achieving a similar function of reducing LLM prior in LVLMs. By editing the LVLMs with \nameshort{}, the lower CHAIR$_S$ (the metric to evaluate hallucination, the lower, the better) indicates that OH is effectively mitigated by \nameshort{}. Therefore, we can consider \nameshort{} as a new way that achieves debiasing from the model parameter aspect.



\subsection{How \nameshort{} works?} 

The analysis is performed for each layer $\ell$, and to avoid the notational burden, we will drop $\ell$ and focus on each layer separately. We also suppose the number of tokens as 1; the generated features can then be calculated as $\vf_i = \vx_i \mW^\text{org}$, where $\mW^\text{org} \in \R^{D\times D_o}$, and $\vx_i  \in \R^{1\times D}$, $\vf_i  \in \R^{1\times D_o}$.


Based on the heuristic in~\cite{uppaal2024detox}, an embedding vector in any transformer layer can be decomposed into interpretable components. We suppose that the generated features $\vf_i$ can be separated into three different elements:
\begin{align}
\label{eq:factor-main}
    \vf_i \Rightarrow  \underbrace{\hat \vf_i \hat \mB }_{\text{truthful contexts}} + \underbrace{\tilde \vf_i \tilde \mB }_{\text{hallucinated biases}} +\underbrace{\vu_i}_{\text{noise}},
\end{align}
where we suppose $\hat \mB \in \R^{\hat k \times D_o}$ contains the truthful directions and $\tilde \mB \in \R^{\tilde k \times D_o}$ contains $\tilde k$ hallucinated directions, and $\vu_i$ represents typical randomness unaccounted for by the statistical model. $\hat \vf_i \in \R^{1\times \hat k}, \tilde \vf_i \in \R^{1\times \tilde k}$ are ``latent factors''. Therefore $\tilde \vf_i \tilde \mB$ can represent the hallucinated features for the $i$-th input.



However, with the $\mW^\text{ed}$ and combining the hypothesize in Eq. (\ref{eq:factor-main}), the edited features can be calculated as
\begin{align} 
\label{eq:factor_de}
    \vf_i^\text{ed}  &= \vx_i (\mI - \mV_k \mV_k^\top)
    \mW^\text{org}  \nonumber \\
    &= \hat \vf_i \hat \mB  + (\tilde \vf_i \tilde \mB -  (\vx_i\mV_k )  (\mV_k^\top\mW^\text{org}) ) + \vu_i.
\end{align}
As the $\mV_k$ gives the best $k$-rank approximation of the difference between the truthful and untruthful features via SVD, our approach can remove the hallucinated features and enhance the truthful answers for accurate responses.

\vspace{-5pt}
\paragraph{Relation to DPO} Following previous studies in~\cite{uppaal2024detox}, we exhibit the conceptual connection between DPO~\cite{rafailov2024direct} and \nameshort{} by studying a simple logistic model $\pi_{\mW}$ for the output given $\vx$ as inputs. The conditional probability can be
\begin{equation} \label{eq:dpo1}\vspace{-5pt}
    \pi_{\mW}(y|\vx_{i}) = Z_{\mW}^{-1} \exp\big( \vo_y^\top \mW \vx_{i} \big)
\vspace{-2pt}
\end{equation}
where $\vo_y$ is the output decoding vector for any $y \in \mathcal{V}$, $Z_{\mW}$ is the normalization factor. Similar expression holds for $\vx_i^+$ and $\vx_i^-$. With some calculations, the gradient with respect to $\mW$ of the DPO loss can be represented as
\begin{align} \label{eq:grad_pi}
&\nabla_{\mW} \mathcal L_{\dpo} = - \frac{\beta}{N}\sum_{i=1}^N \big( \vo_{y_i^+} (\vx_i^+)^\top  -   \vo_{y_i^-} (\vx_i^-)^\top \big) \nonumber \\
 = &- \frac{\beta}{N}\sum_{i=1}^N \big( \underbrace{\vo_{y_i^+} (\vx_i^+ - \vx_i^- )^\top}_{\text{feature difference}} + \underbrace{ (\vo_{y_i^+} - \vo_{y_i^-}) (\vx_i^-)^\top}_{\text{output difference}} \big).
\end{align}
The gradient in (\ref{eq:grad_pi}) contains a feature difference term. Therefore, the gradient update can be interpreted as an attempt to eliminate feature differences to avoid hallucinated responses. For \nameshort{}, it tries to approximate such difference via SVD and also attempts to eliminate it by null space projection, which shows the connection between \nameshort{} and DPO. See Supplementary Materials for more details.





\begin{table*}[htp]
\centering
\setlength{\tabcolsep}{2pt}
\resizebox{\linewidth}{!}{
\begin{tabular}{ l | c c c | c c c | c c c}
\toprule
\multirow{2}{*}{\textbf{Method}} 
&\multicolumn{3}{c|}{LLaVA-1.5}
&\multicolumn{3}{c|}{MiniGPT-4}
&\multicolumn{3}{c}{mPLUG-Owl2} \\
\cmidrule{2-10}
&\textbf{CHAIR}$_S \downarrow$ &\textbf{CHAIR}$_I \downarrow$ &BLEU$\uparrow$ 
&\textbf{CHAIR}$_S \downarrow$ &\textbf{CHAIR}$_I \downarrow$ &BLEU$\uparrow$ 
&\textbf{CHAIR}$_S \downarrow$ &\textbf{CHAIR}$_I \downarrow$ &BLEU$\uparrow$   \\ 
\midrule
Greedy
&$\text{20.40}_{\pm \text{2.80}}$ &$\text{7.08}_{\pm \text{0.33}}$ &$\text{15.72}_{\pm \text{0.10}}$ 
&$\text{32.40}_{\pm \text{2.20}}$ &$\text{12.20}_{\pm \text{0.42}}$ &$\text{14.57}_{\pm \text{0.11}}$ 
&$\text{22.90}_{\pm \text{0.90}}$ &$\text{8.62}_{\pm \text{0.11}}$ &$\text{15.01}_{\pm \text{0.24}}$  \\
Beam Search \cite{freitag2017beam}
&$\text{19.50}_{\pm \text{2.30}}$ &$\text{6.84}_{\pm \text{0.79}}$ &$\text{15.99}_{\pm \text{0.14}}$ 
&$\text{30.10}_{\pm \text{0.30}}$ &$\text{11.87}_{\pm \text{0.37}}$ &$\text{15.35}_{\pm \text{0.24}}$ 
&$\text{20.30}_{\pm \text{0.70}}$ &$\text{7.62}_{\pm \text{0.19}}$ &$\text{15.43}_{\pm \text{0.05}}$  
\\
DoLa \cite{chuang2023dola}
&$\text{20.20}_{\pm \text{2.80}}$ &$\text{6.75}_{\pm \text{0.54}}$ &$\text{15.68}_{\pm \text{0.10}}$ 
&$\text{31.90}_{\pm \text{3.30}}$ &$\text{12.15}_{\pm \text{0.89}}$ &$\text{14.54}_{\pm \text{0.12}}$ 
&$\text{22.40}_{\pm \text{1.80}}$ &$\text{8.36}_{\pm \text{0.04}}$ &$\text{15.13}_{\pm \text{0.21}}$  
\\ 
OPERA \cite{huang2024opera}
&$\text{17.50}_{\pm \text{0.50}}$ &$\text{6.07}_{\pm \text{0.32}}$ &$\text{16.02}_{\pm \text{0.02}}$ 
&$\text{29.70}_{\pm \text{0.30}}$ &$\text{11.96}_{\pm \text{0.29}}$ &$\text{14.82}_{\pm \text{0.05}}$ 
&$\text{20.07}_{\pm \text{2.07}}$ &$\text{7.18}_{\pm \text{0.39}}$ &$\text{15.41}_{\pm \text{0.12}}$ 
\\ 
VCD \cite{leng2024mitigating}
&$\text{20.30}_{\pm \text{1.10}}$ &$\text{7.28}_{\pm \text{0.10}}$ &$\text{14.53}_{\pm \text{0.01}}$ 
&$\text{29.00}_{\pm \text{2.80}}$ &$\text{12.64}_{\pm \text{1.19}}$ &$\text{14.42}_{\pm \text{0.01}}$ 
&$\text{22.80}_{\pm \text{0.80}}$ &$\text{8.68}_{\pm \text{0.17}}$ &$\text{15.14}_{\pm \text{0.13}}$  
\\
Woodpecker \cite{yin2023woodpecker}
&$\text{23.85}_{\pm \text{4.62}}$ &$\text{7.50}_{\pm \text{0.01}}$ &$\text{17.05}_{\pm \text{0.00}}$ 
&$\text{28.87}_{\pm \text{2.20}}$ &$\text{10.20}_{\pm \text{0.85}}$ &$\text{15.30}_{\pm \text{0.01}}$ 
&$\text{26.33}_{\pm \text{1.98}}$ &$\text{8.43}_{\pm \text{0.80}}$ &$\text{16.43}_{\pm \text{0.00}}$  
\\ 
LURE \cite{zhou2023analyzing}
&$\text{19.48}_{\pm \text{2.35}}$ &$\text{6.50}_{\pm \text{0.38}}$ &$\text{15.97}_{\pm \text{0.01}}$   
&$\text{27.88}_{\pm \text{2.25}}$ &$\text{10.20}_{\pm \text{0.85}}$ &$\text{15.03}_{\pm \text{0.11}}$ 
&$\text{21.27}_{\pm \text{0.06}}$ &$\text{7.67}_{\pm \text{0.16}}$ &$\text{15.65}_{\pm \text{0.15}}$ 
\\ 
HALC \cite{chen2024halc}
&$\text{16.90}_{\pm \text{2.10}}$ &$\text{5.72}_{\pm \text{0.55}}$ &$\text{16.02}_{\pm \text{0.04}}$ 
&$\text{25.20}_{\pm \text{2.00}}$ &$\text{9.42}_{\pm \text{0.41}}$ &$\text{14.91}_{\pm \text{0.13}}$ 
&$\text{18.80}_{\pm \text{1.20}}$ &$\text{7.00}_{\pm \text{0.01}}$ &$\text{15.33}_{\pm \text{0.24}}$ 
\\ 
\midrule
\rowcolor{mygray} \textbf{\nameshort} 
&$\text{\textbf{15.20}}_{\pm \text{0.60}}$ &$\text{\textbf{5.30}}_{\pm \text{0.03}}$ &$\text{15.69}_{\pm \text{0.04}}$ 
&$\text{\textbf{21.40}}_{\pm \text{1.00}}$ &$\text{\textbf{8.99}}_{\pm \text{0.36}}$ &$\text{14.81}_{\pm \text{0.06}}$ 
&$\text{\textbf{15.60}}_{\pm \text{1.20}}$ &$\text{\textbf{5.77}}_{\pm \text{0.01}}$ &$\text{15.45}_{\pm \text{0.01}}$ 


\\ \bottomrule
\end{tabular}
}
\vspace{-6pt}
\caption{CHAIR evaluation results on MSCOCO dataset of LVLMs (LLaVA-1.5, MiniGPT-4 and mPLUG-Owl2) with different methods for mitigating OH. Lower $\text{CHAIR}_S$ and $\text{CHAIR}_I$ indicate less OH. Higher BLEU generally represent higher captioning quality. We use 64 as the max token number in this experiment. Bold indicates the best result of all methods.}
\label{tab:chair_results}
\vspace{-8pt}
\end{table*}

\section{Experiments}
\label{sec:exp}
In this section, we conduct experiments to evaluate the effectiveness of our method in mitigating OH. Experimental analyses and a case study are also provided. 

\subsection{Datasets and baselines}

We use the widely used CHAIR~\cite{rohrbach2018object} and POPE~\cite{li2023evaluating}, and randomly sampled 500 images from the validation split of MSCOCO~\cite{lin2014microsoft} for evaluations following previous settings in~\cite{huang2024opera,yin2023woodpecker,chen2024halc}. We repeat the experiments three times for each metric with different random seeds. Moreover, MME~\cite{fu2023mme} and LLaVA-Bench~\cite{liu2023improved} are applied as extensive benchmarks tailored to assess the overall performance after editing. The detailed descriptions and implementations can be found in Supplementary Materials.

\textbf{CHAIR.} CHAIR is a tailored tool created to evaluate the occurrence of OH in an image description by determining the proportion of the mentioned objects that are absent in the ground-truth label set. For metrics, $\text{CHAIR}_S$ measures proportion of the hallucinated sentences over all sentences, and $\text{CHAIR}_I$ measures proportion of the hallucinated objects over all generated objects. Lower scores indicate less OH. We also report BLEU as an assessment of the text generation quality. For implementation, we use LURE~\cite{zhou2023analyzing} as the paired data for \nameshort{}. We prompt all methods with ``\textit{Please describe this image in detail.}''.  

\textbf{POPE.} The POPE dataset presents a streamlined approach to assess object hallucination. With POPE, LVLMs are queried to answer whether a specific object exists in the image. It encompasses three sampling settings: \textit{random, popular, and adversarial}, each distinct in constructing negative samples. Besides the basic evaluation method, we further use offline POPE (OPOPE)~\cite{chen2024halc}, which keeps the object sampling and yes/no query strategy from POPE but replaces the live interactions with offline checks. Specifically, instead of querying the model with ``\textit{Is there a \{\} in the image?}'', where ``\textit{\{\}}'' is the queried object, we first ask the examined LVLM to give its detailed descriptions of the image and then check if the sampled positive/negative objects exist in the captions when computing the OPOPE scores. 


\textbf{MME and LLaVA-Bench.} The Multi-modal Large Language Model Evaluation (MME) benchmark comprises ten perception-related and four cognition-related tasks. We use all tasks to test the edited models comprehensively. Moreover, LLaVA-Bench is a collection of 24 images, where each image is paired with a detailed, manually crafted description and carefully selected questions. Both of these datasets are used to assess the capability of LVLMs to tackle more challenging tasks.


\textbf{LVLM Baselines.} We evaluate the effectiveness of our \nameshort{} on three popular LVLMs, including LLaVA-1.5~\cite{liu2024visual} with Vicuna~\cite{vicuna2023}, MiniGPT-4~\cite{zhu2023minigpt4} with Llama2~\cite{touvron2023llama2} and mPLUG-Owl2~\cite{ye2024mplug}. See supplementary materials for more implementation details.


\begin{table*}[t!]
\centering
\setlength{\tabcolsep}{2pt}
\resizebox{\linewidth}{!}{
\begin{tabular}{ l | c c c | c c c | c c c}
\toprule
\multirow{2}{*}{Method} 
&\multicolumn{3}{c|}{LLaVA-1.5}
&\multicolumn{3}{c|}{MiniGPT-4}
&\multicolumn{3}{c}{mPLUG-Owl2} \\
\cmidrule{2-10}
&\textbf{Accuracy}$\uparrow$ &\textbf{Precision}$\uparrow$  &\textbf{F score}$\uparrow$  
&\textbf{Accuracy}$\uparrow$ &\textbf{Precision}$\uparrow$ &\textbf{F score}$\uparrow$  
&\textbf{Accuracy}$\uparrow$ &\textbf{Precision}$\uparrow$ &\textbf{F score}$\uparrow$     \\ 
\midrule
Greedy
&$\text{79.14}_{\pm \text{0.89}}$ &$\text{91.98}_{\pm \text{0.82}}$ &$\text{90.45}_{\pm \text{0.86}}$ 
&$\text{71.22}_{\pm \text{1.27}}$ &$\text{93.72}_{\pm \text{1.02}}$ &$\text{90.04}_{\pm \text{1.23}}$  
&$\text{76.46}_{\pm \text{0.92}}$ &$\text{88.85}_{\pm \text{1.15}}$ &$\text{87.29}_{\pm \text{1.15}}$  \\
Beam Search \cite{freitag2017beam}
&$\text{79.41}_{\pm \text{0.69}}$ &$\text{92.52}_{\pm \text{0.55}}$ &$\text{90.96}_{\pm \text{0.59}}$ 
&$\text{71.65}_{\pm \text{1.15}}$ &$\text{94.70}_{\pm \text{0.60}}$ &$\text{90.97}_{\pm \text{0.85}}$  
&$\text{76.76}_{\pm \text{1.02}}$ &$\text{90.28}_{\pm \text{0.80}}$ &$\text{88.56}_{\pm \text{0.87}}$  
\\
DoLa \cite{chuang2023dola}
&$\text{78.98}_{\pm \text{0.56}}$ &$\text{91.66}_{\pm \text{0.81}}$ &$\text{90.15}_{\pm \text{0.79}}$ 
&$\text{71.28}_{\pm \text{1.15}}$ &$\text{93.92}_{\pm \text{0.83}}$ &$\text{90.22}_{\pm \text{1.04}}$  
&$\text{76.07}_{\pm \text{1.09}}$ &$\text{88.54}_{\pm \text{1.25}}$ &$\text{86.95}_{\pm \text{1.27}}$  
\\
OPERA \cite{huang2024opera}
&$\text{79.29}_{\pm \text{0.32}}$ &$\text{92.25}_{\pm \text{0.07}}$ &$\text{90.71}_{\pm \text{0.11}}$ 
&$\text{70.48}_{\pm \text{1.63}}$ &$\text{94.41}_{\pm \text{1.11}}$ &$\text{90.66}_{\pm \text{1.42}}$ 
&$\text{75.49}_{\pm \text{1.29}}$ &$\text{91.23}_{\pm \text{1.06}}$ &$\text{89.11}_{\pm \text{1.17}}$ 
\\ 
VCD \cite{leng2024mitigating}
&$\text{78.01}_{\pm \text{0.75}}$ &$\text{91.33}_{\pm \text{0.88}}$ &$\text{89.69}_{\pm \text{0.89}}$ 
&$\text{70.83}_{\pm \text{1.83}}$ &$\text{92.31}_{\pm \text{0.88}}$ &$\text{88.76}_{\pm \text{1.29}}$  
&$\text{75.49}_{\pm \text{1.27}}$ &$\text{88.75}_{\pm \text{1.56}}$ &$\text{87.02}_{\pm \text{1.57}}$  
\\
HALC \cite{chen2024halc}
&$\text{77.87}_{\pm \text{0.22}}$ &$\text{93.17}_{\pm \text{0.39}}$ &$\text{91.25}_{\pm \text{0.38}}$ 
&$\text{71.17}_{\pm \text{0.89}}$ &$\text{94.88}_{\pm \text{0.15}}$ &$\text{90.95}_{\pm \text{0.42}}$  
&$\text{74.93}_{\pm \text{1.09}}$ &$\text{90.20}_{\pm \text{0.90}}$ &$\text{88.12}_{\pm \text{0.99}}$  
\\ 
\midrule
\rowcolor{mygray}\textbf{\nameshort}
&$\text{\textbf{79.52}}_{\pm \text{0.04}}$ &$\text{\textbf{93.46}}_{\pm \text{0.03}}$ &$\text{\textbf{91.79}}_{\pm \text{0.04}}$ 
&$\text{\textbf{71.92}}_{\pm \text{0.39}}$ &$\text{\textbf{95.96}}_{\pm \text{0.65}}$ &$\text{\textbf{92.07}}_{\pm \text{0.65}}$  
&$\text{\textbf{77.09}}_{\pm \text{1.37}}$ &$\text{\textbf{92.83}}_{\pm \text{0.29}}$ &$\text{\textbf{90.80}}_{\pm \text{0.52}}$  
\\ 
\bottomrule
\end{tabular}
}
\vspace{-8pt}
\caption{The OPOPE evaluation results on MSCOCO dataset of LVLMs with different methods for mitigating OH. Higher accuracy, precision, and F score indicate better performance. Bold indicates the best result of all methods.}
\label{tab:pope}
\vspace{-13pt}
\end{table*}

\subsection{Results on CHAIR}

\begin{figure}[htp]
    \centering
    \includegraphics[width=1.03\linewidth]{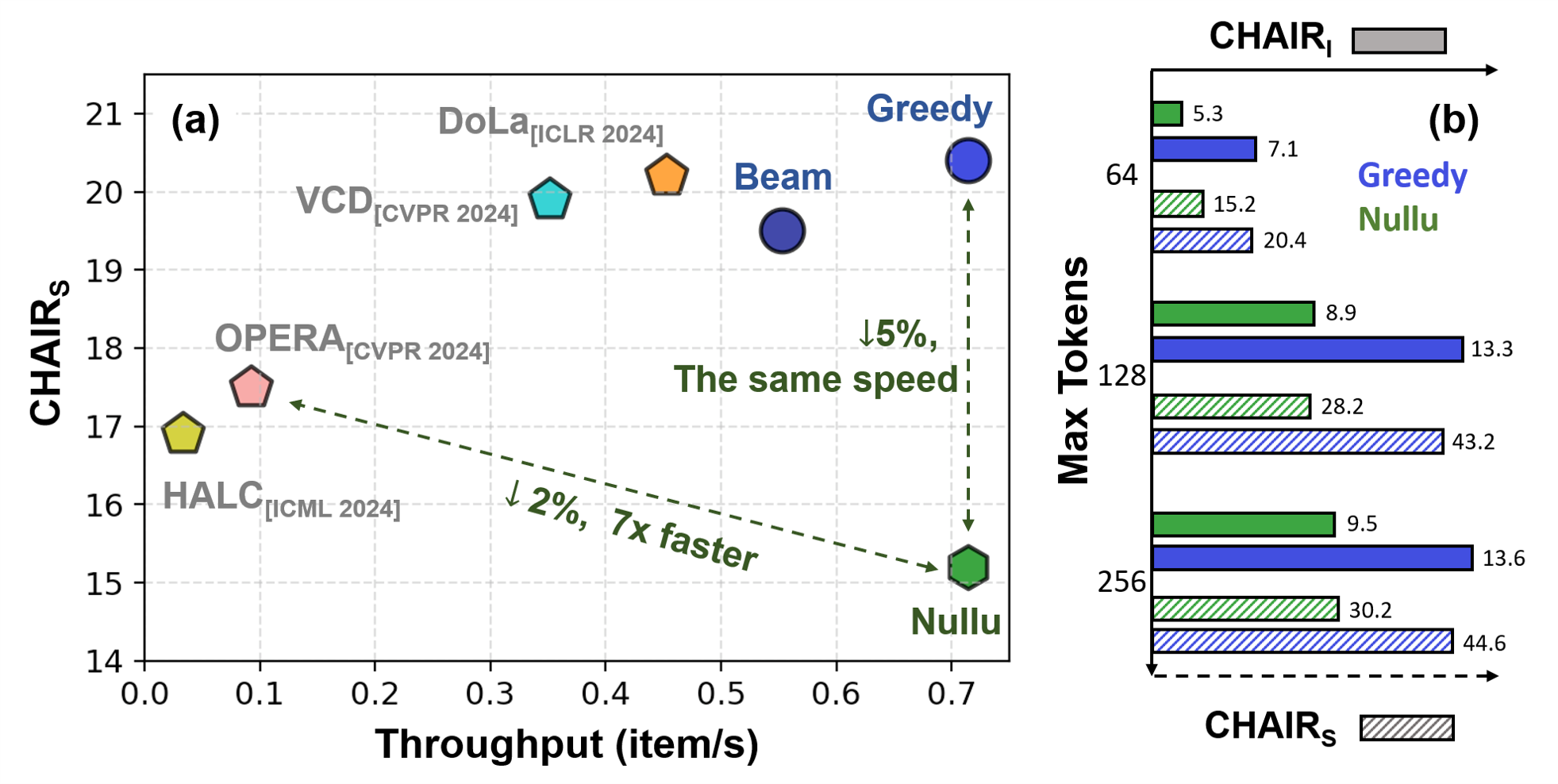} 
    \vspace{-18pt}
    \caption{(a) The throughput (tested on NVIDIA RTX 4090) v.s. CHAIR$_S$. (b) Performance with different max token numbers.}
    \label{fig:speed}
    \vspace{-18pt}
\end{figure}
We provide the experimental results in Table~\ref{tab:chair_results}, from which we see that \nameshort{} consistently outperforms the evaluated methods. For the metric, CHAIR$_\text{S}$ (C$_\text{S}$) can be more critical, since a caption containing many correct objects with one Hallucinated Object is still an error with C$_\text{S}$. A significant improvement in C$_\text{S}$ indicates that Nullu addresses more last hallucinated object. Moreover, although Nullu achieves moderate gain compared to HALC with C$_\text{I}$, it's over 10$\times$ faster than HALC. 

Moreover, we compared our method to a series of baseline models on LLaVA-1.5 with different numbers of max tokens, and the results are shown in Figure~\ref{fig:speed} (b). Results show that more output tokens lead to more hallucinations in the responses. While the proposed \nameshort{} still outperforms the baseline model by maintaining a low ratio of hallucination with longer responses and more generated objects. 

A unique advantage of \nameshort{} is that it can re-parameterize the edited weights to the original model and, therefore, introduces no extra inference costs. However, most post-hoc OH mitigation methods need to modify the final inference procedure by either debiasing the LLM biases with double-inference~\cite{leng2024mitigating} or even using multiple adaptive inference procedures~\cite{chen2024halc} and therefore introduce plenty of extra costs during inference. We test the inference speed of DoLa~\cite{chuang2023dola}, VCD~\cite{leng2024mitigating}, and HALC~\cite{chen2024halc} along with the original LLaVA-1.5 with greedy and beam search decoding strategies. The results are shown in Figure~\ref{fig:speed} (a). Compared to LLaVA-1.5 with greedy decoding strategy, \nameshort{} can decrease the CHAIR$_S$ by over 5\% with the same throughput. Although HALC performs well in mitigating OH, it also introduces many extra inference procedures, which slow the inference speed. Compared to OPERA, \nameshort{} can achieve 7$\times$ faster speed while outperforming OPERA in CHAIR$_S$.

\subsection{Results on POPE}
Experimental results on POPE under the random, popular, and adversarial settings are summarized in the Supplementary Materials. Here, we mainly provide the results using OPOPE metrics, shown in Table~\ref{tab:pope}. All the numbers are averaged results of the three sampling methods. Generally, OPOPE is more challenging than POPE since VLMs first need to describe a whole image and then find a specific object from the response, where the object can often be missing. The results show that \nameshort{} also performs better than all evaluated methods regarding accuracy, precision, and F score metrics. The experimental results demonstrate the effectiveness of \nameshort{} in OH mitigation and show its broad applicability on different open-source LVLMs.

\subsection{Results on MME}
\begin{figure}[htp]
    \centering
    \includegraphics[width=\linewidth]{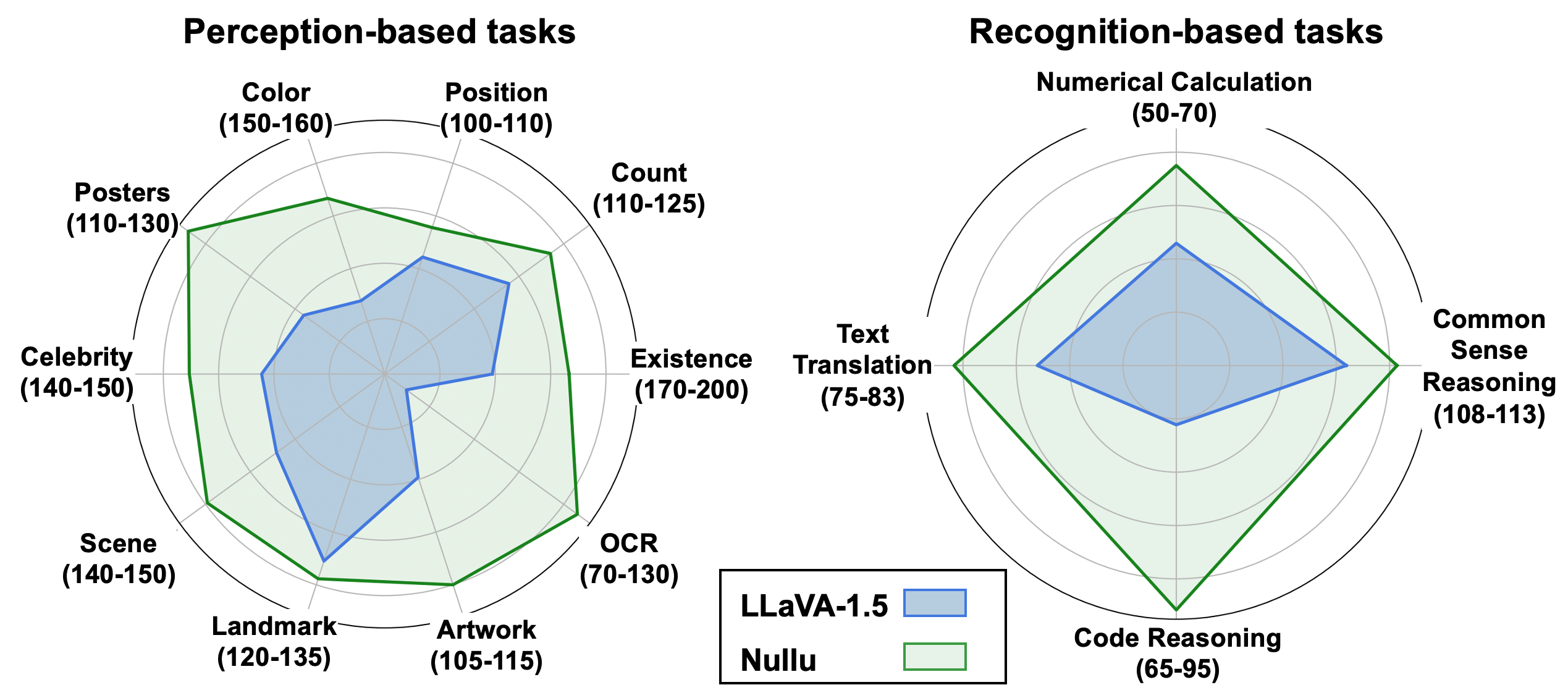}  
    \vspace{-15pt}
    \caption{MME full set evaluation results on LLaVA-1.5.}
    \label{fig:mme}
    \vspace{-10pt}
\end{figure}

We present the results of LLaVA-1.5-7B as a representative to evaluate the general ability of the edited model. From Figure~\ref{fig:mme}, we see that implementing \nameshort{} leads to a consistent enhancement in perception- and recognition-based tasks. Moreover, it is interesting that using \nameshort{} can enhance the LVLM in text-related tasks, such as OCR and Code Sense Reasoning. This may be attributed to the fact that \nameshort{} achieves similar optimization attempts with DPO~\cite{rafailov2024direct}, potentially improving LVLMs' general capacities. See more discussions and numerical results in Supplementary materials.

\subsection{Ablation Studies and Further Analysis}
\begin{figure*}[!t]
    \centering
    \includegraphics[width=\textwidth]{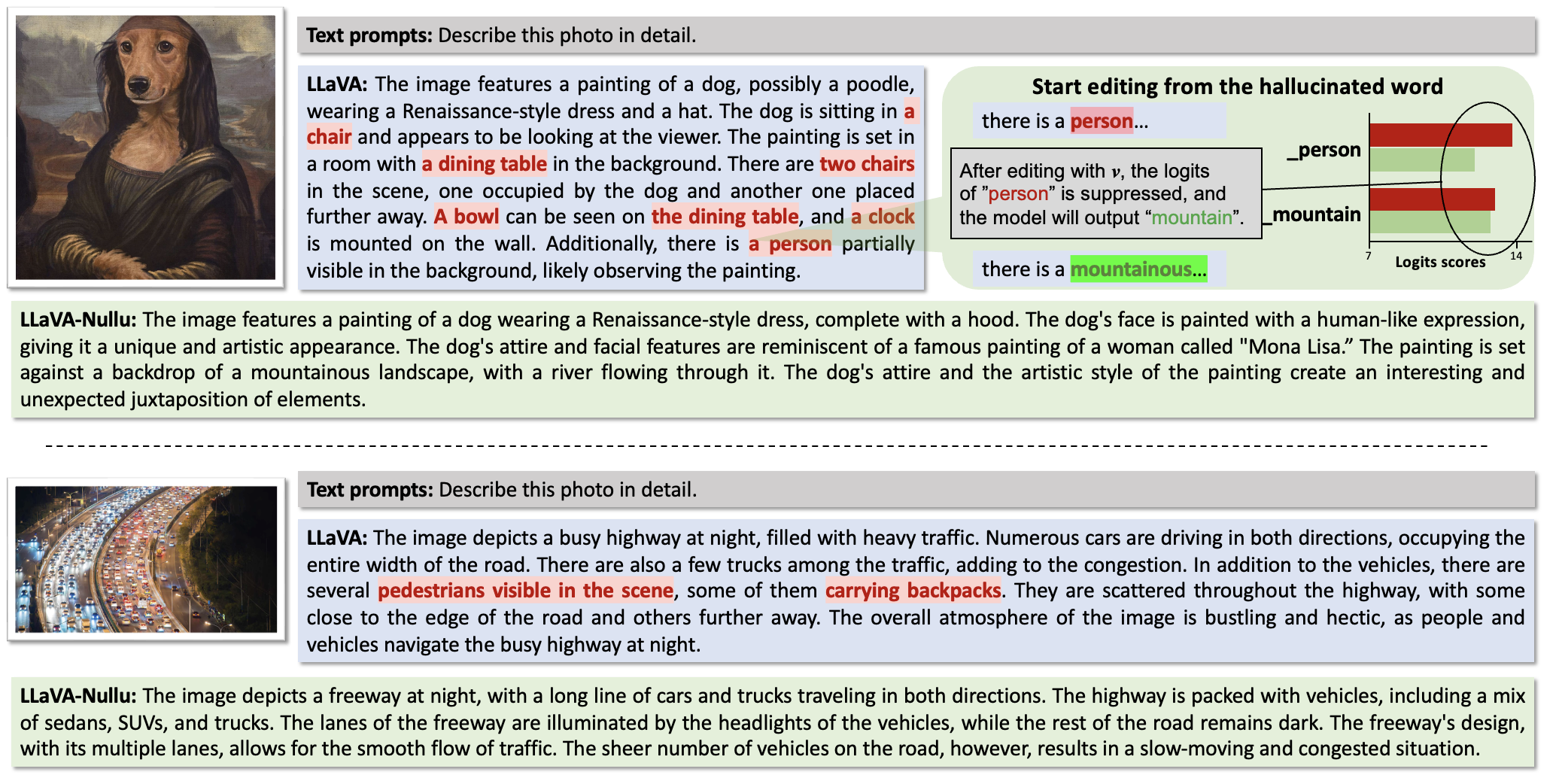}   
    \vspace{-19pt}
    \caption{Illustration of hallucination correction by our proposed \nameshort{} with two samples from LLaVA-Bench using LLaVA-1.5-7B. Hallucinated objects from the original model are highlighted in \textcolor{red}{red}.
    }
    \label{fig:vis}
    \vspace{-10pt}
\end{figure*}

\noindent\textbf{Effects of editing layers $\{\ell\}$ and rank $k$.} \nameshort{} contains two key parameters: the indices of layers needed to be edited $\{\ell\}$ and the selected top-$k$ singular vectors. We test different values of these hyper-parameters with LLaVA-1.5-7B on CHAIR (shown in Table~\ref{tab:lk}). The editing layers can affect overall performance; therefore, we use $\{\ell\}= \{16,..,32\}$ in all experiments. Moreover, we further found that the performance of \nameshort{} with a certain $k$ for building HalluSpace achieves the best performance, indicating that the HalluSpace can be with a low-rank structure. 
\begin{table}[htp]
\vspace{-5pt}
\centering
\resizebox{0.99\linewidth}{!}{%
\begin{tabular}{l|ccc||l|ccc}
\toprule
$\{\ell\}$ & \textbf{C}$_S$ &\textbf{C}$_I $ & \textbf{BLEU}  & k& \textbf{C}$_S$ &\textbf{C}$_I $ &\textbf{BLEU} \\ \midrule
16-32 & \textbf{15.2} & \textbf{5.30} & \textbf{15.7}  &2 &20.2 & 7.22 & 15.3 \\  
24-32 & 16.2 & 5.58 & 15.1 &4 & \textbf{15.2} & \textbf{5.30}   &\textbf{15.7}\\ 
28-32 & 18.2 & 6.27 & 15.1 &8 & 17.8 & 5.92   &15.2 \\
31-32 & 16.4 & 5.59 & 15.1 &16 & 17.2 & 6.27 & 14.8 \\ \bottomrule
\end{tabular}
}
\vspace{-5pt}
\caption{Ablation study of the hyper-parameters in \nameshort{}.}
\label{tab:lk}
\vspace{-10pt}
\end{table}

\begin{figure}[htp]
\vspace{-5pt}
    \centering
    \includegraphics[width=\linewidth]{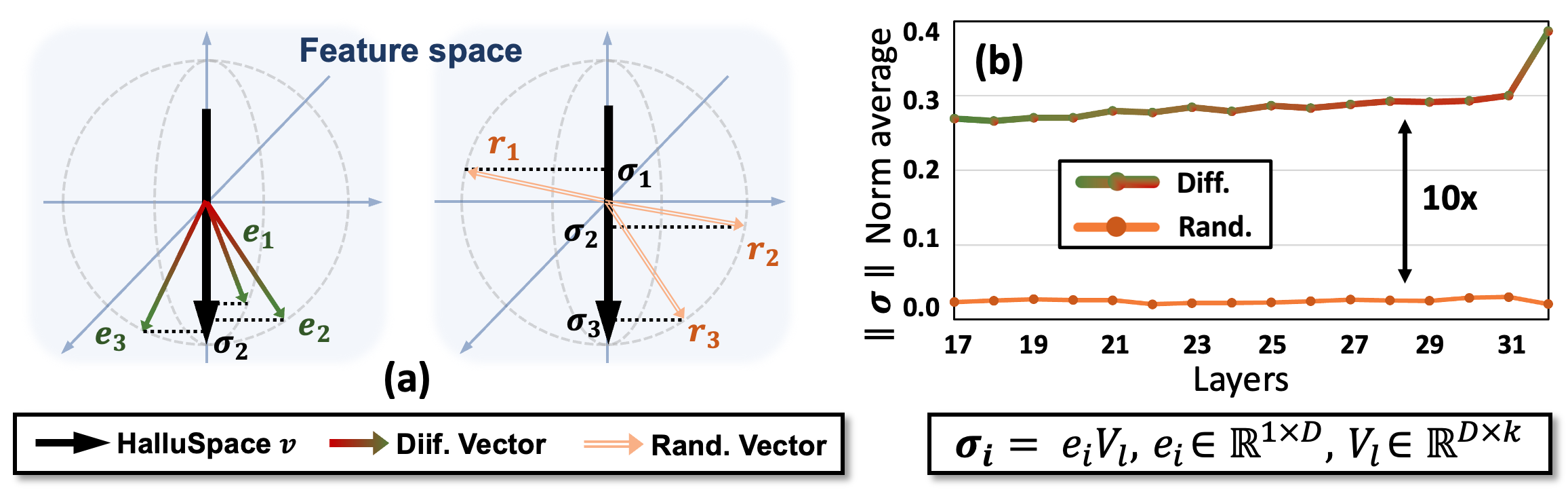}  
    \vspace{-15pt}
    \caption{(a) The illustration of experiments. (b) The mean of $||\sigma||$ of 100 test and random vectors at different layers. }
    \label{fig:f}
    \vspace{-5pt}
\end{figure}

\noindent\textbf{Does HalluSpace represent the hallucination biases?} Ideally, if HalluSpace effectively represents these biases, the difference vectors from test samples should have big projected components when mapped onto HalluSpace. To evaluate this, we select 100 test samples from CHAIR where \nameshort{} successfully mitigates OH issues. We compute difference vectors $\ve_i$ for each sample between the raw and edited LLaVA features. Moreover, we generate 100 random vectors $\vr_i$ as a comparison baseline. All these vectors are normalized to avoid the effects of norms. Moreover, we use $\sigma_i$ to represent the projected components. Figure~\ref{fig:f} (a) shows the distribution of vectors on a normalized sphere. 

Given $\mV_4$ (rank-4), each projected component $\sigma_i$ resides within $\R^4$. We then calculated $\sigma_i$ for all selected and random samples, averaging $||\sigma||$ across samples. Figure~\ref{fig:f} (b) presents these results, showing that the average $||\sigma||$ of difference vectors across layers is significantly larger (10$\times$) than that of random vectors. Since the selected test samples were successfully edited to avoid OH, this evidence indicates that HalluSpace captures directions in the feature space associated with hallucinations in LVLMs.

\begin{table}[htp]
\vspace{-5pt}
\centering
\resizebox{0.8\linewidth}{!}{%
\begin{tabular}{@{}llcc@{}}
\toprule
\textbf{Model}   & \textbf{Method} & Accuracy$\uparrow$ & Detailedness$\uparrow$ \\ \midrule
\multirow{2}{*}{LLaVA-1.5}     
    & Original      & $5.89$            & $5.02$            \\
    & \nameshort{}  & $\textbf{6.53}$   & $\textbf{5.59}$   \\ \midrule
\multirow{2}{*}{MiniGPT-4}      
    & Original      & $4.11$            & $4.07$             \\
    & \nameshort{}  & $\textbf{5.63}$   & $\textbf{4.87}$    \\ \midrule
\multirow{2}{*}{mPLUG-Owl2} 
    & Original      & $5.96$            & $4.52$             \\
    & \nameshort{}  & $\textbf{6.26}$   & $\textbf{4.72}$       \\ \bottomrule
\end{tabular}
}
\vspace{-5pt}
\caption{Results of GPT-4V-aided evaluation on LLaVA-Bench following the setting in~\cite{leng2024mitigating}. Both metrics are on a scale of 10.}
\vspace{-10pt}
\label{tab:gpt4v}
\end{table}
\noindent\textbf{GPT-4V Aided Evaluation on LLaVA-Bench.}
Following~\cite{leng2024mitigating,yin2023woodpecker}, we leverage LLaVA-Bench~\cite{liu2023improved} to qualitatively evaluate the performance using GPT-4V Aided Evaluation. The prompt used for evaluation and an evaluation case are provided in Supplementary Materials. Results in Table~\ref{tab:gpt4v} show consistent improvements using \nameshort{} for each model, demonstrating our method's effectiveness in both OH mitigation and general model ability improvement.

\noindent\textbf{Case Study on LLaVA-Bench}
Figure~\ref{fig:vis} demonstrates two case studies on LLaVA-1.5-7B given identical prompts and images. The results show that the original LLaVA wrongly describes the object ``\textit{chair}'', then it hallucinates the frequently co-occurring objects such as ``\textit{\textcolor{red}{dining table}}'' and ``\textit{\textcolor{red}{bowl}}''. A similar issue also happens in the second example, where the model wrongly generates ``\textit{\textcolor{red}{pedestrians}}'' due to the possible descriptions of the ``\textit{traffic}''. In contrast, implementing \nameshort{} can mitigate these hallucination issues effectively. We further test \nameshort{} in an online setting, where we edited the features with the learned $\mV$ before a hallucinated word, and we found that null space projection effectively fixed the hallucinated word ``\textcolor{red}{person}'' to a correct word ``mountain''. More results can be found in our Supplementary Materials.



%

\vspace{-10pt}
\section{Conclusion and Future Work}
\vspace{-5pt}
In this paper, we proposed a novel method, \nameshort{}, to address OH in modern LVLMs. \nameshort{} extracts HalluSpaces, which are low-rank subspaces of the differences between truthful and hallucinated features, and further edits the LVLMs weights to mitigate OH. Empirical results show that using the edited model weights reduces the priors in the LLMs, which has been proven as an essential factor for OH. Experiments demonstrate that \nameshort{} can significantly mitigate OH with no extra inference cost, making it much more efficient than many existing methods and maintaining strong performance in general LVLM benchmarks.

Considering the connection between \nameshort{} and DPO, \nameshort{} can potentially be an alternative but effective way for safety alignment, providing a more efficient and straightforward way to improve LVLMs trustworthiness.

\section*{Acknowledgments}
This work is supported in part by the National Key Research and Development Program of China (2023YFE0209800) and the National Natural Science Foundation of China (62206215, U24B20185, 62161160337 and 62376210).

{
    \small
    \bibliographystyle{ieeenat_fullname}
    \bibliography{main}
}


\clearpage

\section{The derivation of null space}
\label{nullspace}
Here, we give the details about obtaining the null space of the $\vv$. We want to proof that: any vector $\vz \in \mathbb{R}^D$ in the null space of the $\vv \in \mathbb{R}^D$, $(\mI - \vv\vv^\top)$, is orthogonal to the vector $\vv$. Namely, we have $\vv^\top \vz =0$, where $\vv$ is the vector with norm 1. $\mI$ is the identity matrix with the size of $\mathbb{R}^{D\times D}$. We can write the $\vz$ as 
\begin{align}
\label{eq:null1}
   \vz = (\mI - \vv\vv^\top) \vm, \ \ \ \forall \vm \in \mathbb{R}^D.
\end{align}
Then we have
\begin{align}
\label{eq:null2}
   \vv^\top  \vz &= \vv^\top (\mI - \vv\vv^\top) \vm = (\vv^\top - (\vv^\top\vv) \vv^\top)\vm, \\
   &= (\vv^\top -  \vv^\top)\vm = 0,  \ \ \forall \vm \in \mathbb{R}^D.
\end{align}
Therefore, $(\mI - \vv\vv^\top)$ is the null space of $\vv$.

\section{Decoding information in HalluSpace}
\begin{table}[h]
  \centering
    \resizebox{1\linewidth}{!}{
    \begin{tabular}{c|llllllll}
    \toprule
    \textbf{layer} & \multicolumn{7}{c}{\textbf{Top Tokens}} \\
    \midrule
    16    & dynamic  & either  & further  & above  & \textcolor{red}{background}& floor  & tables & ...  \\
    17    & another & notable  & left  & later & others  & most  & tables & ... \\
    18    & nearby  & notable  & either  & tables  & group  & optional  & others & ... \\
    19    & notable  & \textcolor{red}{middle}  & either  & diverse  & \textcolor{red}{background}  & overall  & concentr & ... \\
    20    & notable  & left  & nearby  & either  & \textcolor{red}{background}  & center  & \textcolor{red}{middle} & ... \\
    21    & \textcolor{red}{middle}  & another  & left  & bottom  & top   & left  & right & ...\\
    22    & left  & position  & \textcolor{red}{middle}  & another  & right & \textcolor{red}{background}  & top & ... \\
    23    & notable  & left  & position  & nearby  & left  & another  & bottom  & ...\\
    24    & position  & notable  & various  & various  & \textcolor{red}{middle}  & \textcolor{red}{background}  & above & ... \\
    25    & position  & towards  & left  & nearby  & right  & another  & bottom & ... \\
    26    & in    & position  & towards  & positions  & left  & nearby  & engaged & ... \\
    27    & in    & position  & towards  & closer  & nearby  & right  & \textcolor{red}{background} & ... \\
    28    & in    & the   & position  & closer  & \textcolor{red}{background}  & nearby  & right  & ...\\
    29    & in    & position  & closer  & towards & nearby  & \textcolor{red}{background}  & a & ...\\
    30    & in    & closer  & nearby  & right  & left  & another  & top & ... \\
    31    & closer  & close  & position  & bottom  & another  & left  & top & ... \\
    \bottomrule
    \end{tabular}
    }
  \caption{LLaVA-1.5-7B, top-rank-4, each singular vector of the matrix is interpreted by identifying the top 10 tokens it represents. We use the output embedding vector $e_j$ to find top-scoring tokens $j\in \mathcal{V}$ for maximizing $\langle v_i, e_j \rangle$. Tokens have been censored for readability.}
  \label{tab:llavawordtop4}
\end{table}

As we state in the main paper, to explore the information behind the learned vectors $\vv_i \in \R^D$, a common approach is to decode the embeddings with $\mO  \vv \in \R^{|\mathcal{V}|}$, where $\mO = [\vo_1, \ldots, \vo_{|\mathcal{V}|}]^\top \in \R^{|\mathcal{V}| \times D}$ and $\mathcal{V}$ denotes the vocabulary. We then sort $\mO \vv$ in ascending order, find the top-$m$ indices, and use the corresponding words to interpret $\vv$. 

Using \nameshort{}, we can extract the $\mV$ at different layers and then decode it via $\mO \vv$ to explore the internal information behind $\mV$. We provide the decoding results in Table.~\ref{tab:llavawordtop4}. Moreover, we select the words with the most frequency in the output of LVLM with distorted images. For a more straightforward interpretation, we directly selected the words in Table.~\ref{tab:llavawordtop4} to see the frequency of words before and after \nameshort{} to see if the LLM biases are mitigated.

\section{Theoretical Analysis: How \nameshort{} works?}

\subsection{Factor component analysis}
The analysis is performed for each layer $\ell$, and to avoid the notational burden, we will drop $\ell$ and focus on each layer separately. We use the same notations with these in the main paper. Based on the heuristic in~\cite{uppaal2024detox}, an embedding vector in any transformer layer can be decomposed into interpretable components. We suppose that the generated features $\vf_i$ can be separated into three different elements:
\begin{align}
\label{eq:factor-main-sup}
    \vf_i \Rightarrow  \underbrace{\hat \vf_i \hat \mB }_{\text{truthful contexts}} + \underbrace{\tilde \vf_i \tilde \mB }_{\text{hallucinated biases}} +\underbrace{\vu_i}_{\text{noise}}.
\end{align}

Therefore, give positive and negative samples as input, we have
\begin{align}
\label{eq:factor-main12}
    \vf_i^+ =&  \underbrace{\hat \vf_i \hat \mB }_{\text{truthful contexts}} + \underbrace{\tilde \vf_i \tilde \mB }_{\text{hallucinated biases}} &+\underbrace{\vu_i^+}_{\text{noise}}.  \\
    \vf_i^- =&  \underbrace{\hat \vf_i \hat \mB }_{\text{truthful contexts}}  &+\underbrace{\vu_i^-}_{\text{noise}},
\end{align}
based on which we have 
\begin{align}
\label{eq:factor-main3}
    \mE = \vf_i^+ - \vf_i^- = \tilde \vf_i \tilde \mB + (\vu_i^+ - \vu_i^-). 
\end{align}
The noise can be approximated to 0 on average of the whole data. The top-$k$ singular vectors span exactly the same subspace of $\tilde \mB$, which can be the HalluSpace in our paper. Moreover, SVD is also efficient since SVD gives the best low-rank approximation of $\mE$. Thus, our approach can be viewed as an approximate recovery of the latent subspace for hallucination semantics.

\subsection{Connections to DPO}
In this subsection, we try to establish the conceptual connection between DPO~\cite{rafailov2024direct} and the proposed \nameshort{}. Our study is mainly based on the theoretical analysis in~\cite{uppaal2024detox}, where a simple logistic model for the output token given the (continuing) prompt is used. In the following parts, we will drop $\ell$ and focus on each layer separately to avoid notational burden.

Although the proposed \nameshort{} is designed for LVLMs, we mainly study its LLM parts, since our weight editing is mainly conducted on this part. Therefore, in this section, we use the term input to denote the extracted features $\vx$, containing both the visual features processed by the previous visual encoder, and the text prompts projected into the embedding space. Given $\vx$ with hallucinated response $y^+$ and truthful response $y^-$, where the corresponding embedding features denoted as $\vx, \vy^+,\vy^-$ respectively, DPO optimizes the loss

\begin{small}
\begin{align}
    \Ls_{\dpo}(\pi_\vtheta;\pi_{\text{ref}}) = &-\E_{(x, y^+, y^-)\sim \cal D} [\log\sigma (\beta\log\dfrac{\pi_\theta(\vy^+|\vx)}{\pi_{\text{ref}}(\vy^+|\vx)} \nonumber \\ 
    &-\beta\log\dfrac{\pi_\theta(\vy^-|\vx)}{\pi_{\text{ref}}(\vy^-|\vx)})],
\end{align}
\end{small}
where, $\pi_{\text{ref}}$ corresponds to the reference (or base) probability model generating output $y$ given $x$, $\pi_\vtheta$ is the new probability model (parametrized by $\vtheta$), $\sigma$ is the logistic function with $\sigma(z)=(1+\exp(-z))^{-1}$, and $\beta>0$ is a hyperparameter. The gradient of the loss $\Ls_\dpo$ with respect to $\theta$ at initialization $\pi_\theta=\pi_{\text{ref}}$ equals
\begin{small}
\begin{align}\label{eq_appndx:DPO_first_gradient}
    &\nabla_\vtheta \cal L_\dpo(\pi_\vtheta;\pi_{\text{ref}})\mid_{\pi_\vtheta=\pi_{\text{ref}}} \nonumber \\
    & = -\beta \E_{(x,y^+,y^-)\sim \cal D} [\nabla_\vtheta \log\pi(\vy^+|\vx) - \nabla_\vtheta\log\pi(\vy^-|\vx)]\mid_{\pi_\theta=\pi_{\text{ref}}}.
\end{align}
\end{small}

Let $\cal V$ denote the vocabulary. We start with an input $x$ (including both textual and visual features) and produce $M$ next-token predictions $y_1,\cdots, y_M\in\cal V$ sequentially.
Suppose the model sequentially predicts token $y_m$ given $x_m:=(x,y_1,\cdots, y_{m-1})$ for each $1\leq m\leq M$, and let $\vx_m$ denote the encoding of input $x_m$. We assume a logistic model generating each continuation $y_m$ given $x_m$, 
\begin{align}
\label{dpo3}
    \pi_\vtheta(y_m|x_m) \equiv \pi_\vW(y_m|x_m) = Z_{m,\vW}^{-1}\exp\left( \vo_{y_m}^\top \vW\vx_m\right).
\end{align}
Here, $\vo_{y_m}$ is the classification vector which we use to get the final word prediction, $\vW$ is a weight matrix and $Z_{m,\vW}$ is the normalizing constant:
\begin{align*}
    Z_{m,\vW} &= \sum_{y\in\cal V}\exp\left(\vo_{y}^\top \vW\vx_m\right).
\end{align*}

For the results in Eq. (\ref{dpo3}), we have assumed for simplicity that the classification is performed with linearly transformed encoding $\vW\vx_m$ instead of the more common non-linear transformations in the transformer architecture. And the output probability is given by the logistic model, based on which we can obtain the joint probability of observing the entire continuation $y=(y_1,\cdots, y_M)$ given the starting input $x$ as
\begin{small}
\begin{align*}
    \pi_\vtheta(y|x) &\equiv \pi_\mW(y|x) = \prod_{m=1}^M \pi_\mW(y_m|x_m) \nonumber \\
    & = Z_\mW^{-1}\exp\left(\sum_{m=1}^M \vo_{y_m}^\top \vW\vx_m\right),
\end{align*}
\end{small}
where $Z_\mW = \prod_{m=1}^M Z_{m,\mW}$.
We denote by $x_m^\pm$, $\vx_m^\pm$ and $\vo_{y_m}^\pm$ the positive/negative inputs, the corresponding embedding and classification vector for the positive/negative continuation respectively. Plugging this into (\ref{eq_appndx:DPO_first_gradient}), the first step DPO update has gradient
\begin{small}
\begin{align}
    &\nabla_\mW\Ls_\dpo(\pi_\mW;\pi_{\text{ref}})|_{\pi_\mW = \pi_{\text{ref}}}  \nonumber \\
    &= -\beta\E_{(x,y^+,y^-)\sim\cal D}\left[\sum_{m=1}^M \left(\vo_{y_m}^+(\vx_m^+)^\top - \vo_{\vy_m}^{-}(\vx_m^-)^\top\right)\right].
\end{align}
\end{small}
Note that the the normalization factors $Z_{m, \mW}$ (and hence $Z_\mW$) are omitted when we take the difference of the gradients of the log-probabilities. With $N$ pairs of inputs in $\cal D$, and we consider the case $M=1$, the DPO gradient will be an average over all the pairs:
\begin{small}
\begin{align}
\label{dpo_final}
    &\nabla_\mW\Ls_\dpo(\pi_\mW;\pi_{\text{ref}})|_{\pi_\mW = \pi_{\text{ref}}} 
    = -\dfrac{\beta}{N}\sum_{i=1}^N(\vo_{y_{i}}^+(\vx_{i}^+)^\top - \vo_{y_{i}}^{-}(\vx_{i}^-)^\top ),
\end{align}
\end{small}
where the extra index $i$ mean $i$-th sample pairs. The Eq.(\ref{dpo_final}) is the formulation (7) in our main paper, which is 
\begin{align}
&\nabla_{\mW} \mathcal L_{\dpo} = - \frac{\beta}{N}\sum_{i=1}^N \big( \vo_{y_i^+} (\vx_i^+)^\top  -   \vo_{y_i^-} (\vx_i^-)^\top \big) \nonumber \\
 = &- \frac{\beta}{N}\sum_{i=1}^N \big( \underbrace{\vo_{y_i^+} (\vx_i^+ - \vx_i^- )^\top}_{\text{feature difference}} + \underbrace{ (\vo_{y_i^+} - \vo_{y_i^-}) (\vx_i^-)^\top}_{\text{output difference}} \big).
\end{align}

The gradient contains a feature difference term. Therefore, the gradient update can be interpreted as an attempt to eliminate feature differences to avoid hallucinated responses. For \nameshort{}, it tries to approximate such difference via SVD and also attempts to eliminate it by null space projection, which shows the connection between \nameshort{} and DPO.

\section{Implementation Details of LVLMs}\label{app:exp}
This section details the implementation of the evaluated LVLMs and the methods used for OH mitigation. The overall experimental setup is summarized in Table~\ref{tab:maxtokens_overall}. Unlike the standard greedy method, which selects the most probable token at each decoding step, beam search maintains a fixed number of candidate sequences (beams) per step, ranking them based on the accumulated probability scores of the previous tokens $(y_{<t})$. In our experiments, the beam search method uses a \textit{num-beams} setting of 3, specifying the number of candidate sequences retained at each step.
We use the default code for implementation of these two baselines in HuggingFace Transformers Repository\cite{wolf2020transformers}.\footnote{\url{https://huggingface.co/docs/transformers}}
\begin{table}[H]
\centering
\begin{tabular}{l|c}
\hline
\textbf{Parameters} & \textbf{Value} \\ \hline
Do-sample   & False \\ \hline
Num-beams (for beam search)  & $3$ \\ \hline
Maximum New Tokens (CHAIR) & $64$  \\ \hline
Maximum New Tokens (POPE)  & $64$  \\ \hline
Maximum New Tokens (MME)  & $64$ \\ \hline
Maximum New Tokens (OPOPE)  & $256$  \\ \hline
Maximum New Tokens (LLaVA-Bench)  & $1024$ \\ \hline
\end{tabular}
\caption{Hyper-parameters for LVLMs.}
\label{tab:maxtokens_overall}
\end{table}
%

The complete hyper-parameters for Nullu across different models in our experiments are as follows. Specifically, there are three major hyper-parameters that can be actively adjusted to optimize Nullu's effectiveness across different models:

\begin{enumerate}
    \item Editing Layers $\ell$: For all models, the editing layers are specified by $\ell \in \text{range}(16,32)$.
    \item The Selected Top-\textit{k} singular vector: The number of top-\textit{k} singular vectors selected varies by model. We use the value 4 for LLaVA-1.5 on both CHAIR and POPE. Similarly, we use 8 for MiniGPT-4 on the evaluated two datasets. For mPLUG-Owl2, we use 32 on CHAIR and 16 on POPE.
    \item Num-beams: This parameter also differs across models. It is set to 3 for both LLaVA-1.5 and MiniGPT-4, while for mPLUG-Owl2, it is set to 1.
\end{enumerate}
For the comparison of Nullu with SOTAs methods specifically designed for OH mitigation, the evaluation code is built based on the public repository of HALC~\cite{chen2024halc}\footnote{\url{https://github.com/BillChan226/HALC}}.
Specifically, the hyper-parameters for HALC, VCD~\cite{leng2024mitigating}, DoLa~\cite{chuang2023dola} and OPERA~\cite{huang2024opera} are reported in Table~\ref{tab:hyperparameter_halc}, Table~\ref{tab:hyperparameter_vcd},  Table~\ref{tab:hyperparameter_dola} and Table~\ref{tab:hyperparameter_opera}, respectively. For each baseline, we follow the official implementation and use the pre-trained models and configurations from their respective repositories to reproduce the reported results.

\begin{table}[H]
\centering
\begin{tabular}{l|c}
\hline
\textbf{Parameters} & \textbf{Value} \\ \hline
Amplification Factor $\alpha$ & $0.05$  \\ \hline
JSD Buffer Size $m$ & $6$  \\ \hline
Beam Size & $1$ \\ \hline
FOV Sampling & Exponential Expansion  \\ \hline
Number of Sampled FOVs $n$  & $4$  \\ \hline
Exponential Growth Factor $\lambda$ & 0.6\\ \hline
Adaptive Plausibility Threshold & $0.1$ \\
\hline
\end{tabular}
\caption{HALC Hyperparameter Settings}
\label{tab:hyperparameter_halc}
\end{table}
\begin{table}[H]
\centering
\begin{tabular}{l|c}
\hline
\textbf{Parameters} & \textbf{Value} \\ \hline
Amplification Factor $\alpha$ & $1$  \\ \hline
Adaptive Plausibility Threshold $\beta$ & $0.1$\\\hline
Diffusion Noise Step & $500$\\
\hline
\end{tabular}
\caption{VCD Hyperparameter Settings}
\label{tab:hyperparameter_vcd}
\end{table}
\begin{table}[H]
\centering
\begin{tabular}{l|c}
\hline
\textbf{Parameters} & \textbf{Value} \\ \hline
Repetition Penalty $\theta$ & $1.2$  \\ \hline
Adaptive Plausibility Threshold $\beta$ & $0.1$\\ \hline
Pre-mature Layers & $[0, 2 \cdots, 32]$\\
\hline
\end{tabular}
\caption{DoLa Hyperparameter Settings}
\label{tab:hyperparameter_dola}
\end{table}
\begin{table}[H]
\centering
\begin{tabular}{l|c}
\hline
\textbf{Parameters} & \textbf{Value} \\ \hline
Self-attention Weights Scale Factor $\theta$ & $50$  \\ \hline
Attending Retrospection Threshold & $15$\\\hline
Beam Size & $3$\\
\hline
Penalty Weights & $1$\\
\hline
\end{tabular}
\caption{OPERA Hyperparameter Settings}
\label{tab:hyperparameter_opera}
\end{table}

\section{POPE Settings and Additional Results}

Polling-based Object Probing Evaluation (POPE)~\cite{li2023evaluating}, presents a streamlined approach to assess object hallucination. POPE interacts directly with the examined LVLM, which distinguishes it from CHAIR. Within this benchmark, LVLMs are queried to answer if a specific object exists in the given image. The ratio between queries probing existent objects and non-existent objects is balanced (i.e.,$50$\% vs. $50$\%). It encompasses three sampling settings: \textit{random, popular, and adversarial}, each distinct in constructing negative samples. In the \textit{random} setting, objects absent from the image are chosen randomly. The \textit{popular} setting selects missing objects from a high-frequency pool, while in the \textit{adversarial} setting, co-occurring objects not present in the image are prioritized. We use the POPE benchmark aggregating data from MSCOCO~\cite{lin2014microsoft}. For each experiment, we select $500$ images under each sampling setting and generate $6$ questions per image. The evaluation pivots on four key metrics: Accuracy, Precision, Recall, and the F1 score.


\subsection{POPE Results}
\begin{table*}[htp]
\centering
\resizebox{0.8\linewidth}{!}{%
    \begin{tabular}{lcllll|c}
    \toprule
    {\textbf{Setting}} & {\textbf{Model}} & \textbf{Method} & \textbf{Accuracy} & \textbf{Precision} & \textbf{Recall} & \textbf{$\text{F}_\text{1}$ Score} \\
    \midrule
    \multirow{6}[2]{*}{\textit{Random}} & \multirow{6}[2]{*}{MiniGPT4} & Greedy & 64.33  & 58.66  & 97.13  & 73.14  \\
          &       & Beam Search  & 62.10  & 57.15  & 96.67  & 71.84  \\
          &       & DoLa  & 64.27  & 58.82  & 95.10  & 72.68   \\
          &       & VCD   & 57.90  & 55.69  & 77.27  & 64.73  \\
          &       & HALC  & 64.87  & 59.04  & 97.13  & 73.44  \\
          &       & \cellcolor{black!10}Nullu & \cellcolor{black!10}77.23  & \cellcolor{black!10}76.54  & \cellcolor{black!10}78.53  & \cellcolor{black!10}\textbf{77.53} \\
\cmidrule{2-7}    \multirow{6}[2]{*}{\textit{Popular}} & \multirow{6}[2]{*}{MiniGPT4} & Greedy & 56.63  & 53.66  & 97.13  & 69.13  \\
          &       & Beam Search  & 56.47  & 53.58  & 96.67  & 68.95  \\
          &       & DoLa  & 56.58  & 53.72  & 95.10  & 68.65   \\
          &       & VCD   & 55.30  & 53.59  & 79.20  & 63.92  \\
          &       & HALC  & 57.00  & 53.88  & 97.13  & 69.31  \\
          &       &\cellcolor{black!10} Nullu & \cellcolor{black!10}70.13  & \cellcolor{black!10}67.24  & \cellcolor{black!10}78.53  & \cellcolor{black!10}\textbf{72.45} \\
\cmidrule{2-7}    \multirow{6}[2]{*}{\textit{Adversarial}} & \multirow{6}[2]{*}{MiniGPT4} & Greedy & 55.17  & 52.81  & 97.13  & 68.42  \\
          &       & Beam Search  & 55.50  & 53.02  & 96.67  & 68.48  \\
          &       & DoLa  & 55.85  & 53.28  & 95.10  & 68.29   \\
          &       & VCD   & 52.90  & 51.99  & 75.60  & 61.61  \\
          &       & HALC  & 55.53  & 53.02  & 97.13  & 68.60  \\
          &       & \cellcolor{black!10}Nullu & \cellcolor{black!10}66.70  & \cellcolor{black!10}63.50  & \cellcolor{black!10}78.53  & \cellcolor{black!10}\textbf{70.22} \\
    \bottomrule
    \end{tabular}%
}
\caption{POPE results with random, popular and adversarial samplings compared to existing OH mitigation methods.}
\label{tab:pope_result}
\end{table*}


We conduct the comparison between the raw LVLMs and the one implemented with \nameshort{} on POPE and provide the results in Table~\ref{tab:pope_raw_edit}. 

\begin{table*}[]
  \centering
  \resizebox{0.8\linewidth}{!}{%
    \begin{tabular}{ccllllc}
    \toprule
    \multicolumn{1}{l}{\textbf{Setting}} & \multicolumn{1}{l}{\textbf{Model}} & \textbf{Method} & \textbf{Accuracy} & \textbf{Precision} & \textbf{Recall} & \textbf{$\textbf{F}_\textbf{1}$ Score} \\
    \midrule
    \multirow{6}[6]{*}{\textit{random}} & \multirow{2}[2]{*}{LLaVA-1.5} & Original & 88.98  & 88.65  & 89.43  & 89.03  \\
          &       & \cellcolor{black!10}Nullu & \cellcolor{black!10}89.45  & \cellcolor{black!10}91.41  & \cellcolor{black!10}87.10  & \cellcolor{black!10}\textbf{89.20} \\
\cmidrule{2-7}          & \multirow{2}[2]{*}{MiniGPT4} & Original & 64.33  & 58.66  & 97.13  & 73.14  \\
          &       & \cellcolor{black!10}Nullu & \cellcolor{black!10}77.23  & \cellcolor{black!10}76.54  & \cellcolor{black!10}78.53  & \cellcolor{black!10}\textbf{77.53} \\
\cmidrule{2-7}          & \multirow{2}[2]{*}{mPLUG-Owl2} & Original & 81.83  & 77.80  & 89.07  & 83.06  \\
          &       & \cellcolor{black!10}Nullu & \cellcolor{black!10}83.33  & \cellcolor{black!10}79.10  & \cellcolor{black!10}90.60  & \cellcolor{black!10}\textbf{84.46} \\
    \midrule
    \multirow{6}[6]{*}{\textit{popular}} & \multirow{2}[2]{*}{LLaVA-1.5} & Original & 84.58  & 81.61  & 89.43  & 85.32  \\
          &       & \cellcolor{black!10}Nullu & \cellcolor{black!10}85.37  & \cellcolor{black!10}84.25  & \cellcolor{black!10}87.10  & \cellcolor{black!10}\textbf{85.63} \\
\cmidrule{2-7}          & \multirow{2}[2]{*}{MiniGPT4} & Original & 56.63  & 53.66  & 97.13  & 69.13  \\
          &       & \cellcolor{black!10}Nullu & \cellcolor{black!10}70.13  & \cellcolor{black!10}67.24  & \cellcolor{black!10}78.53  & \cellcolor{black!10}\textbf{72.45} \\
\cmidrule{2-7}          & \multirow{2}[2]{*}{mPLUG-Owl2} & Original & 75.77  & 70.35  & 89.07  & 78.61  \\
          &       & \cellcolor{black!10}Nullu & \cellcolor{black!10}77.47  & \cellcolor{black!10}71.75  & \cellcolor{black!10}90.60  & \cellcolor{black!10}\textbf{80.08} \\
    \midrule
    \multirow{6}[6]{*}{\textit{adversarial}} & \multirow{2}[2]{*}{LLaVA-1.5} & Original & 77.97  & 72.79  & 89.43  & 80.24  \\
          &       & \cellcolor{black!10}Nullu & \cellcolor{black!10}79.40  & \cellcolor{black!10}75.51  & \cellcolor{black!10}87.10  & \cellcolor{black!10}\textbf{80.88} \\
\cmidrule{2-7}          & \multirow{2}[2]{*}{MiniGPT4} & Original & 55.17  & 52.81  & 97.13  & 68.42  \\
          &       & \cellcolor{black!10}Nullu & \cellcolor{black!10}66.70  & \cellcolor{black!10}63.50  & \cellcolor{black!10}78.53  & \cellcolor{black!10}\textbf{70.22} \\
\cmidrule{2-7}          & \multirow{2}[2]{*}{mPLUG-Owl2} & Original & 72.77  & 67.17  & 89.07  & 76.58  \\
          &       & \cellcolor{black!10}Nullu & \cellcolor{black!10}74.03  & \cellcolor{black!10}68.05  & \cellcolor{black!10}90.60  & \cellcolor{black!10}\textbf{77.72} \\
    \bottomrule
    \end{tabular}}
    \caption{Results on POPE. Original denotes direct sampling for LVLMs, whereas \nameshort{} refers to edit the model with the proposed method.}
  \label{tab:pope_raw_edit}%
\end{table*}%

We also tested different OH methods on MiniGPT-4 and provided the results in Table~\ref{tab:pope_result}. The results show that \nameshort{} outperforms all other methods by a significant margin regarding the accuracy and F1 score across all three types of POPE VQA tasks (random, popular, adversarial). Our experiments show that the MiniGPT-4 tends to provide the answer with ``yes'', which leads to a high recall ratio for most tested OH methods. However, the Precision of these methods is generally lower than 60\%, resulting in a lower F1 score. However, \nameshort{} significantly improves the Precision of the MiniGPT-4, resulting in a noticeable improvement in the F1 score. Moreover, we also see that VCD also has a lower recall, indicating that the LLM bias of Mini-GPT makes the model tend to provide the answer with ``yes'' when responding.


\begin{table*}[htp]
\centering
\addtolength{\tabcolsep}{2pt}  
\renewcommand{\arraystretch}{0.95}  
\fontsize{8.5pt}{8.5pt}\selectfont
\resizebox{0.81\linewidth}{!}{
    \begin{tabular}{@{}ccllll|c}
    \toprule
    {\textbf{Setting}} & {\textbf{Model}} & \textbf{Method} & \textbf{Accuracy} & \textbf{Precision} & \textbf{Recall} & \textbf{F Score} \\
    \midrule
    \multirow{21}[6]{*}{\textit{Random}} & \multirow{7}[2]{*}{LLaVA-1.5} & Greedy & 81.52  & 98.41  & 64.07  & 96.42  \\
          &       & Beam Search & 81.67  & 98.67  & 64.20  & \textbf{96.67}  \\
          &       & DoLa  & 81.38  & 98.11  & 64.00  & 96.14  \\
          &       & OPERA & 81.62  & 98.57  & 64.17  & 96.58  \\
          &       & VCD   & 80.57  & 98.41  & 62.13  & 96.25  \\
          &       & HALC  & 79.58  & 98.21  & 60.27  & 95.89  \\
          &       & \cellcolor{black!10} Nullu & \cellcolor{black!10}81.18  & \cellcolor{black!10}98.05  & \cellcolor{black!10}63.63  & \cellcolor{black!10}96.05  \\
\cmidrule{2-7}          & \multirow{7}[2]{*}{MiniGPT-4} & Greedy & 72.42 & 98.49 & 45.53 & 94.25 \\
          &       & Beam Search & 72.65 & 98.70 & 45.90 & 94.51 \\
          &       & DoLa  & 72.45 & 98.58 & 45.57 & 94.34 \\
          &       & OPERA & 72.57 & 98.77 & 45.70 & 94.52 \\
          &       & VCD   & 72.35 & 98.19 & 45.53 & 93.97 \\
          &       & HALC  & 72.08 & 98.62 & 44.80 & 94.25 \\
          &       & \cellcolor{black!10}Nullu & \cellcolor{black!10}72.68 & \cellcolor{black!10}99.06 & \cellcolor{black!10}45.80 & \cellcolor{black!10}\textbf{94.82} \\
\cmidrule{2-7}          & \multirow{7}[2]{*}{mPLUG-Owl2} & Greedy & 79.45  & 97.74  & 60.30  & 95.46  \\
          &       & Beam Search & 79.45  & 97.52  & 60.43  & 95.27  \\
          &       & DoLa  & 78.33  & 97.60  & 58.10  & 95.09  \\
          &       & OPERA & 78.31  & 97.73  & 57.96  & 95.21  \\
          &       & VCD   & 78.19  & 98.23  & 57.42  & 95.61  \\
          &       & HALC  & 77.83  & 97.72  & 57.00  & 95.10  \\
          &       & \cellcolor{black!10}Nullu & \cellcolor{black!10}80.30  & \cellcolor{black!10}98.40  & \cellcolor{black!10}61.60  & \cellcolor{black!10}\textbf{96.19}  \\
    \midrule
    \multirow{21}[6]{*}{\textit{Popular}} & \multirow{7}[2]{*}{LLaVA-1.5} & Greedy & 78.93  & 91.17  & 64.07  & 89.71  \\
          &       & Beam Search & 79.30  & 91.98  & 64.20  & 90.47  \\
          &       & DoLa  & 78.72  & 90.69  & 64.00  & 89.26  \\
          &       & OPERA & 79.22  & 91.80  & 64.17  & 90.30  \\
          &       & VCD   & 77.57  & 89.87  & 62.13  & 88.35  \\
          &       & HALC  & 77.47  & 91.87  & 60.27  & 90.05  \\
          &       & \cellcolor{black!10}Nullu & \cellcolor{black!10}79.80  & \cellcolor{black!10}94.06  & \cellcolor{black!10}63.63  & \cellcolor{black!10}\textbf{92.36}  \\
\cmidrule{2-7}          & \multirow{7}[2]{*}{MiniGPT-4} & Greedy & 70.80 & 92.01 & 45.53 & 88.53 \\
          &       & Beam Search & 71.32 & 93.35 & 45.90 & 89.77 \\
          &       & DoLa  & 70.90 & 92.33 & 45.57 & 88.82 \\
          &       & OPERA & 71.10 & 92.82 & 45.70 & 89.27 \\
          &       & VCD   & 70.33 & 90.30 & 45.53 & 86.98 \\
          &       & HALC  & 70.92 & 93.80 & 44.80 & 90.00 \\
          &       & \cellcolor{black!10}Nullu & \cellcolor{black!10}71.97 & \cellcolor{black!10}96.08 & \cellcolor{black!10}45.80 & \cellcolor{black!10}\textbf{92.19} \\
\cmidrule{2-7}          & \multirow{7}[2]{*}{mPLUG-Owl2} & Greedy & 76.00  & 87.90  & 60.30  & 86.38  \\
          &       & Beam Search & 75.90  & 87.50  & 60.43  & 86.02  \\
          &       & DoLa  & 75.20  & 88.36  & 58.10  & 86.60  \\
          &       & OPERA & 75.02  & 88.06  & 57.96  & 86.33  \\
          &       & VCD   & 74.86  & 88.16  & 57.42  & 86.37  \\
          &       & HALC  & 75.77  & 91.34  & 57.00  & 89.26  \\
          &       & \cellcolor{black!10}Nullu & \cellcolor{black!10}78.20  & \cellcolor{black!10}92.22  & \cellcolor{black!10}61.60  & \cellcolor{black!10}\textbf{90.49}  \\
    \midrule
    \multirow{21}[6]{*}{\textit{Adversarial}} & \multirow{7}[2]{*}{LLaVA-1.5} & Greedy & 76.97  & 86.36  & 64.07  & 85.22  \\
          &       & Beam Search & 77.27  & 86.92  & 64.20  & 85.75  \\
          &       & DoLa  & 76.85  & 86.18  & 64.00  & 85.05  \\
          &       & OPERA & 77.03  & 86.40  & 64.17  & 85.26  \\
          &       & VCD   & 75.88  & 85.71  & 62.13  & 84.48  \\
          &       & HALC  & 76.57  & 89.44  & 60.27  & \textbf{87.80}  \\
          &       & \cellcolor{black!10}Nullu & \cellcolor{black!10}77.58  & \cellcolor{black!10}88.27  & \cellcolor{black!10}63.63  & \cellcolor{black!10}86.98  \\
\cmidrule{2-7}          & \multirow{7}[2]{*}{MiniGPT-4} & Greedy & 70.43 & 90.65 & 45.53 & 87.32 \\
          &       & Beam Search & 70.98 & 92.06 & 45.90 & 88.63 \\
          &       & DoLa  & 70.50 & 90.85 & 45.57 & 87.50 \\
          &       & OPERA & 70.78 & 91.63 & 45.70 & 88.21 \\
          &       & VCD   & 69.82 & 88.43 & 45.53 & 85.32 \\
          &       & HALC  & 70.52 & 92.22 & 44.80 & 88.60 \\
          &       & \cellcolor{black!10}Nullu & \cellcolor{black!10}71.10 & \cellcolor{black!10}92.73 & \cellcolor{black!10}45.80 & \cellcolor{black!10}\textbf{89.21} \\
\cmidrule{2-7}          & \multirow{7}[2]{*}{mPLUG-Owl2} & Greedy & 74.23  & 83.58  & 60.30  & 82.36  \\
          &       & Beam Search & 73.78  & 82.51  & 60.43  & 81.37  \\
          &       & DoLa  & 73.52  & 83.98  & 58.10  & 82.55  \\
          &       & OPERA & 73.17  & 83.45  & 57.96  & 82.06  \\
          &       & VCD   & 72.85  & 83.01  & 57.42  & 81.61  \\
          &       & HALC  & 74.02  & 86.41  & 57.00  & 84.72  \\
          &       & \cellcolor{black!10}Nullu & \cellcolor{black!10}76.90  & \cellcolor{black!10}88.76  & \cellcolor{black!10}61.60  & \cellcolor{black!10}\textbf{87.28}  \\
    \bottomrule
    \end{tabular}%
}
\caption{Detailed OPOPE results with random, popular and adversarial samplings.}
\label{tab:opope_results}
\end{table*}

\subsection{OPOPE results}

While this interaction is not problematic for evaluating decoding-based baselines, it limits the applicability of POPE to post-hoc OH mitigation methods. This direct interaction also creates greater instability when the examined LVLM is based on smaller language backbones, such as LLaMA-7B, which has less robust chat capabilities. To address these issues, offline POPE (OPOPE) was introduced in HALC~\cite{chen2024halc}, where a comparison is made between this approach and other effective decoding methods.

Since OPOPE evaluates directly based on the caption generated for each image, it follows the caption generation procedure from CHAIR but differs in the subsequent metric calculation. When computing the OPOPE scores, we follow the processing procedure of CHAIR while adopting POPE’s metric calculation methodology.

For every sampled 500 images in the validation split of MSCOCO. The captions generated by the models are tokenized separately and then each word is singularized. Subsequently, the words are mapped to MSCOCO objects using the synonym and double-word lists provided in \cite{rohrbach2018object}.

Next, three hallucination test object lists are constructed following the sampling strategies proposed in the POPE method. We refer detailed explanations of the different options to its original paper\cite{li2023evaluating}. Each list contains six objects, with a 1:1 ratio of ground-truth to nonexistent objects to ensure label balance. These lists are originally used to generate polling questions based on the template ``\textit{Is there a/an \{\} in the image?}'' in \cite{li2023evaluating}.

After obtaining the objects set from the generated captions and the three test objects list, we assess whether the captions include the ground-truth or nonexistent objects. The comparison results are used to compute scores as the score of the corresponding sampling strategy setting.

The primary metric in OPOPE is adjusted to enable more reliable comparisons. Since offline evaluations are less likely to include the exact hallucinated objects in descriptions, false negatives (FNs) and the resulting recall become less reliable. To address this, and in line with HALC, we adopt F-beta as the main metric for OPOPE instead of F-1, reducing the emphasis on FNs. Specifically, the F-beta score is defined as: $F_\beta = (1+\beta^2)\cdot(\text{precision}\cdot\text{recall})/(\beta^2\cdot\text{precision}+\text{recall})$, where $\beta = 0.2$ is used throughout our experiments following~\cite{chen2024halc}.

The detailed and comprehensive evaluation results under each sampling strategy incorporating OPOPE are presented in Table~\ref{tab:opope_results}. From the results, we see that our method achieve 7 best results (denoted by bold) in 9 comparisons, which again demonstrates the effectiveness of our method.

\section{MME Numerical Results}

\begin{table*}[]
  \centering
  \resizebox{\linewidth}{!}{
    \addtolength{\tabcolsep}{1pt}  
    \renewcommand{\arraystretch}{1}  
    \begin{tabular}{clccccc||ll}
\toprule
\multicolumn{1}{l}{Model} & Method & \textit{Existence} & \textit{Count} & \textit{Position} & \textit{Color} & \textit{Posters} & &\textit{\textbf{Perception Total}} \\
\midrule   
\multirow{2}{*}{LLaVA-1.5} 
    & Original & $181.67_{\pm 2.36}$  & $118.33_{\pm 12.47}$  & $104.44_{\pm 10.39}$  & $152.78_{\pm 5.67}$  & $117.23_{\pm 4.79}$  & \multirow{2}{*}{Original}  &\multirow{2}{*}{$1246.36_{\pm 5.79}$}   \\
    & Nullu & $\textbf{190.00}_{\pm 4.08}$  & $\textbf{121.11}_{\pm 7.74}$  & $\textbf{105.56}_{\pm 8.20}$  & $\textbf{156.67}_{\pm 9.81}$  & $\textbf{127.55}_{\pm 4.20}$ & & \\
    \cmidrule{1-7} 
    \multicolumn{1}{l}{Model} & Method & \textit{Celebrity} & \textit{Scene} & \textit{Landmark} & \textit{Artwork} & \textit{OCR} &  \multirow{3}{*}{\nameshort{}}& \multirow{3}{*}{$\textbf{1330.71}_{\pm 19.77}$}\\
    \cmidrule{1-7} 
    \multirow{2}{*}{LLaVA-1.5} 
    & Original & $111.67_{\pm 3.90}$  & $144.83_{\pm 1.50}$  & $130.65_{\pm 5.26}$  & $108.92_{\pm 2.99}$  & $75.83_{\pm 5.89}$  & & \\
    & Nullu & $\textbf{115.59}_{\pm 6.60}$  & $\textbf{147.92}_{\pm 1.36}$  & $\textbf{131.66}_{\pm 1.09}$  & $\textbf{113.00}_{\pm 2.07}$  & $\textbf{121.67}_{\pm 8.25}$  &  \\
    \bottomrule
    \end{tabular}%
    }
    \caption{Results on all MME perception-related tasks.}
    \label{tab:mme_preception}
\end{table*}

\begin{table*}[]
    \centering
  \resizebox{0.8\linewidth}{!}{
    \addtolength{\tabcolsep}{2pt}  
    \renewcommand{\arraystretch}{1}  
    \begin{tabular}{rlcccc|c}
    \toprule
    \multicolumn{1}{l}{Model} & Method & \multicolumn{1}{c}{\textit{\begin{tabular}[c]{@{}c@{}}Common Sense\\ Reasoning\end{tabular}}} & \multicolumn{1}{c}{\textit{\begin{tabular}[c]{@{}c@{}}Numerical\\ Calculation\end{tabular}}} & \multicolumn{1}{c}{\textit{\begin{tabular}[c]{@{}c@{}}Text\\ Translation\end{tabular}}} & \multicolumn{1}{c|}{\textit{\begin{tabular}[c]{@{}c@{}}Code\\ Reasoning\end{tabular}}} & \multicolumn{1}{c}{\textit{\textbf{\begin{tabular}[c]{@{}c@{}}Recognition\\ Total\end{tabular}}}} \\
    \midrule
    \multirow{2}{*}{LLaVA-1.5}  & Original & $111.19_{\pm 4.68}$  & $59.17_{\pm 7.73}$  & $79.17_{\pm 8.25}$  & $71.67_{\pm 11.24}$  & $321.19_{\pm 2.15}$  \\
          & Nullu & $\textbf{112.14}_{\pm 3.55}$  & $\textbf{65.00}_{\pm 16.20}$  & $\textbf{81.67}_{\pm 7.73}$  & $\textbf{92.50}_{\pm 15.94}$  & $\textbf{351.31}_{\pm 2.61}$  \\
    \bottomrule
    \end{tabular}%
  }
    \caption{Results on all MME recognition-related tasks.}
    \label{tab:mme_recognition}
\end{table*}%

In Table~\ref{tab:mme_preception}, we present the performance of the edited LLaVA-1.5 baselines on the perception-related tasks of the MME benchmark. 

The baselines demonstrate consistent performance patterns, with Nullu uniformly improving the perceptual competencies of the LVLM model. Specifically, the edited model shows improvement for tasks typically used to estimate hallucination capability \cite{chen2024halc}, including color, existence, count, and position. Furthermore, likely due to Nullu’s effect in alleviating language priors, the model exhibits enhancements across all tasks, particularly in OCR, achieving an additional 84.35-point improvement in the total score.
Furthermore, Table~\ref{tab:mme_recognition} showcases the performances on recognition-related tasks within the MME benchmark. 
The results suggest that implementing Nullu while mitigating hallucination issues and enhancing perceptual capabilities does not compromise the inherent reasoning abilities of LVLM. This is evident from the consistent overall recognition scores, which indicate that the model's fidelity remains unaffected by the intervention. Nullu significantly surpasses the original model, demonstrating a comprehensive performance improvement in reducing OH while maintaining generation quality.

\section{Analysis about HalluSpace}
\begin{figure}[htp]
    \centering
    \includegraphics[width=\linewidth]{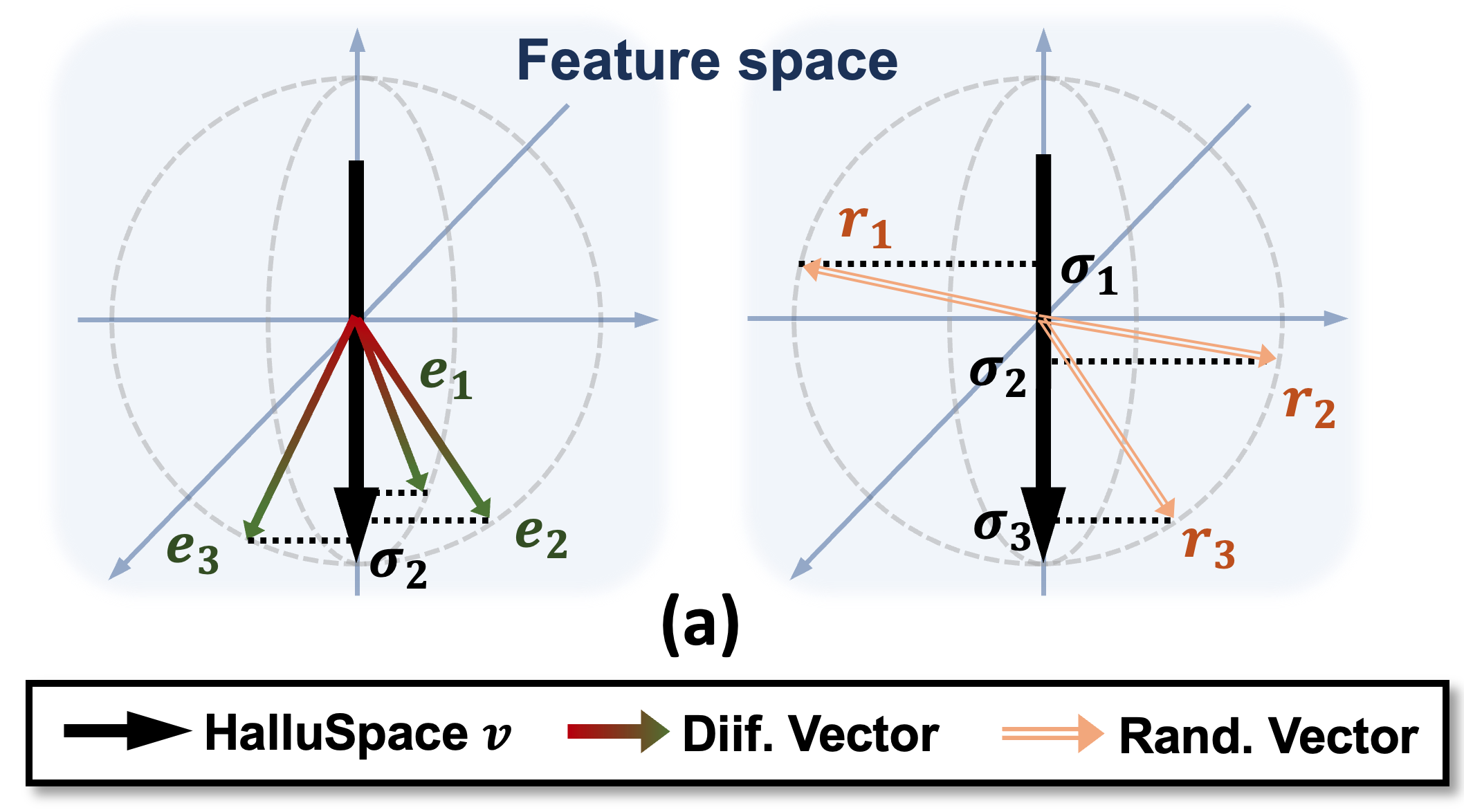}  
    \caption{The illustration of difference vectors and random vectors in the feature space.}
    \label{fig:sm_f1}
\end{figure}

This section provides a more comprehensive study about the question \textbf{Does HalluSpace represent the hallucination biases?}. In other words, can the HalluSpace learned from the prepared hallucinated pairs adequately represent the true OH during the test? Ideally, if HalluSpace effectively represents these biases, the difference vectors from test samples with few OH issues after editing should have larger projected components when mapped onto HalluSpace than these random ones. Indeed, if the HalluSpace represents the OH problematic direction, the aforementioned difference vectors from test samples should gather together around this direction. This is further illustrated in Figure.~\ref{fig:sm_f1}.

To evaluate this, we select 100 test samples from CHAIR where \nameshort{} successfully mitigates OH issues. We compute difference vectors $\ve_i$ for each sample between the raw and edited LLaVA features. Moreover, we generate 100 random vectors $\vr_i$ as a comparison baseline. All these vectors are normalized to avoid the effects of norms. Moreover, we use $\sigma_i$ to represent the projected components. Figure.~\ref{fig:sm_f1} shows the distribution of vectors on a normalized sphere. 

Given $\mV_4$ (rank-4), each projected component $\sigma_i$ resides within $\R^4$. We then calculated $\sigma_i = \ve_i\mV_4$ for all selected samples and random vectors ($\sigma_i = \vr_i\mV_4$), averaging $||\sigma||$ across samples. The results is provided in Table~\ref{tab:smf2}. The table shows that the average $||\sigma||$ of difference vectors across layers is significantly larger (10$\times$) than that of random vectors. Since the selected test samples were successfully edited to avoid OH, this evidence suggests that HalluSpace captures directions in the feature space associated with hallucination and inaccuracies in LVLM responses. 

\begin{table*}[htp]
  \centering
  \resizebox{\linewidth}{!}{
    \addtolength{\tabcolsep}{1pt}  
    \renewcommand{\arraystretch}{1}  
    \begin{tabular}{lllllllllllllllll}
    \toprule
Layers	&	17 	&	18 	&	19 	&	20 	&	21 	&	22 	&	23 	&	24 	&	25 	&	26 	&	27 	&	28 	&	29 	&	30 	&	31 	&	32 	\\ \midrule
Diff.	&	0.269 	&	0.266 	&	0.270 	&	0.270 	&	0.279 	&	0.278 	&	0.284 	&	0.279 	&	0.287 	&	0.283 	&	0.288 	&	0.292 	&	0.291 	&	0.293 	&	0.300 	&	0.386 	\\
Rand	&	0.023 	&	0.025 	&	0.027 	&	0.026 	&	0.026 	&	0.021 	&	0.022 	&	0.022 	&	0.023 	&	0.025 	&	0.027 	&	0.026 	&	0.025 	&	0.029 	&	0.030 	&	0.021 \\
    \bottomrule
    \end{tabular}%
    }
    \caption{Norm average of difference vectors and random vectors at different layers in the LVLM.}
    \label{tab:smf2}
\end{table*}

\section{LLaVA-Bench}

\subsection{Prompt for GPT-4V Aided Evaluation}
As we leverage LLaVA-Bench~\cite{liu2023improved} to qualitatively evaluate the overall performance using GPT-4V Aided Evaluation\footnote{\url{https://openai.com/research/gpt-4v-system-card}}, in this section, we main describe the prompt used for evaluation. The assessments using GPT-4V are based on the accuracy and level of detail in the responses generated by LVLMs, following the approach described~in~\cite{leng2024mitigating}. The specific prompt structure is detailed in Table~\ref{tab:prompt_evaluation}. During the evaluation, we collect the responses from two different LVLMs and then use the responses to replace the ``\{\textit{Response}\}'' in the prompt, which is then sent to GPT-4V for scoring. Next, we analyze the GPT-4V outputs to assess the accuracy and detailedness of the LVLMs' responses. We further provide an evaluation example in Table~\ref{tab:prompt_example} to further illustrate this process.

\begin{table*}[h!]\centering
\begin{minipage}{0.95\textwidth}
\centering
\begin{tcolorbox} 
    \centering
   
      \small
    \begin{tabular}{p{0.95\textwidth}} \hline \\
   \textbf{Description:} \\    
   
   AI that scores image description accuracy and detailedness.

   \\ \midrule

   \textbf{Instructions:} \\   
   
You are an AI designed to evaluate and score the performance of two AI assistants in describing a given image. Your primary focus is on the accuracy and detailedness of their descriptions. You will assess the accuracy by checking for hallucinations - any part of the description that is inconsistent with the image content. For detailedness, you will consider how rich the response is in necessary details, excluding any hallucinated parts. You will provide scores on a scale from 1 to 10 for each assistant separately, based on these criteria. After scoring, you will offer an explanation for your evaluation, ensuring it is free from bias and not influenced by the order of presentation of the responses.
\\ \\
Input format: \\ \\
\lbrack{}Assistant 1\rbrack{}\\
 \{Response 1\}  \\
\lbrack{}End of Assistant 1\rbrack{} \\
\\
\lbrack{}Assistant 2\rbrack{} \\
 \{Response 2\}\\
\lbrack{}End of Assistant 2\rbrack{} \\
\\
Output format:\\
\\
Accuracy:\\
Scores of the two answers:\\
Reason:\\
\\
Detailedness:\\
Scores of the two answers:\\
Reason:\\ \\

\bottomrule
    \end{tabular}
\end{tcolorbox}
\caption{The configuration to build an image-description evaluator with GPT-4V}
    \label{tab:prompt_evaluation}
\end{minipage}
\end{table*}

\begin{table*}[h!]
\centering
\begin{minipage}{0.95\textwidth}
\centering
\begin{tcolorbox}
    \centering
    \small
    \definecolor{customblue}{rgb}{0.17, 0.32, 0.61}
    \begin{tabular}{p{0.95\textwidth}}
    {\normalsize \textcolor{customblue}{\textbf{Input:}}} \\    
    \includegraphics[width=0.2\textwidth]{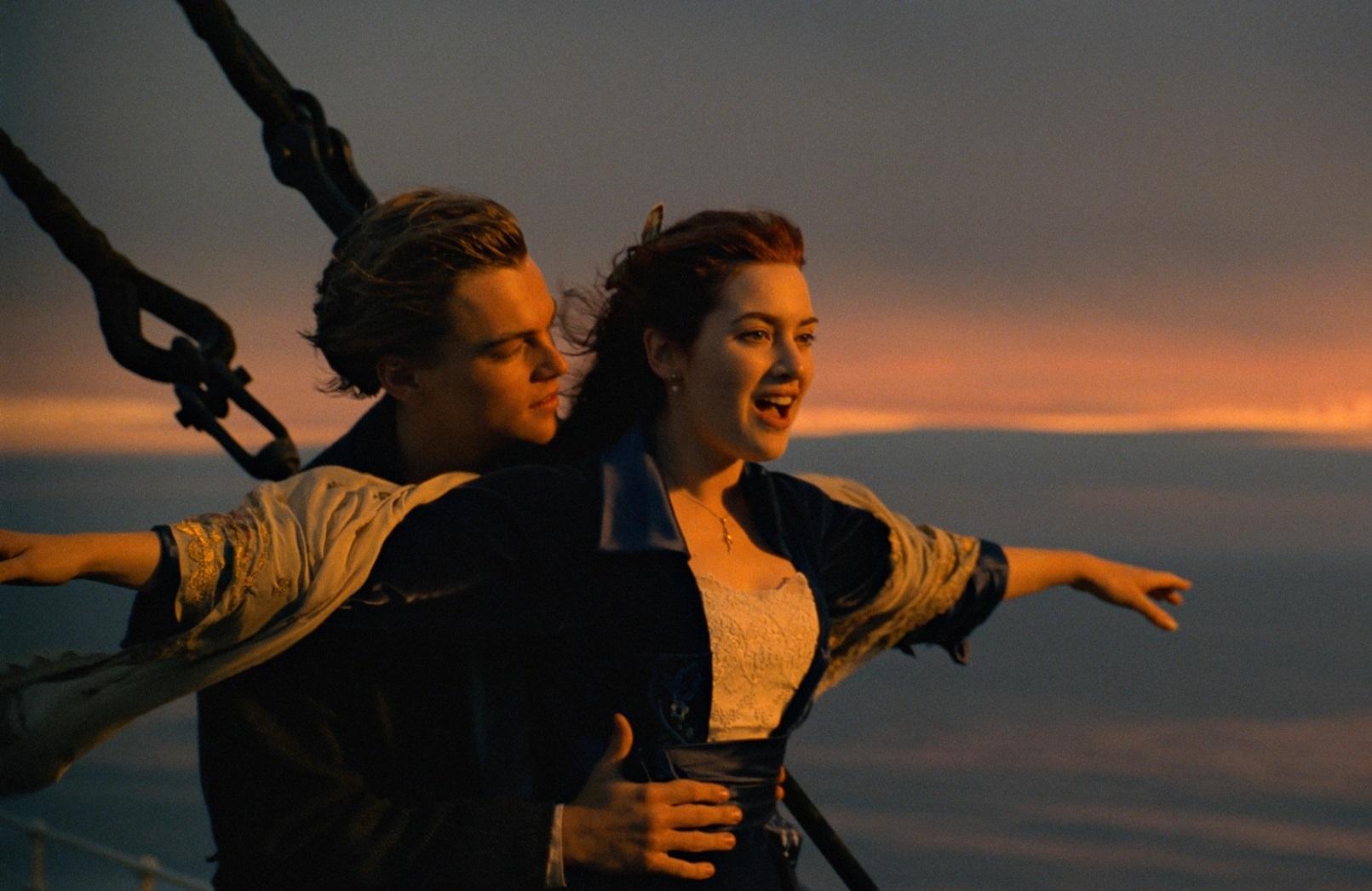} \\ 
    
    {\normalsize \textcolor{customblue}{\textbf{\lbrack{}Assistant 1\rbrack{}}}} \\ 
    In the context of the movie, this scene is significant as it captures a romantic moment between the two main characters, Jack and Rose, who are sharing a passionate kiss while being on a boat in the ocean. This scene is crucial as it highlights the emotional connection between the two characters and adds to the overall narrative of the movie. The kiss takes place during a storm, which adds to the intensity and drama of the scene, further emphasizing the strong bond between the two characters and their willingness to face challenges together. 
     \\
    {\normalsize \textcolor{customblue}{\textbf{\lbrack{}End of Assistant 1\rbrack{}}}} \\ \\
    {\normalsize \textcolor{customblue}{\textbf{\lbrack{}Assistant 2\rbrack{}}}} \\ 
    In the context of the movie "Titanic," this scene of a man and a woman (Rose DeWitt Bukater and Jack Dawson) hanging onto a rope while being on a boat or ship at sea signifies their romantic connection and the beginning of their relationship. It's a pivotal moment in their story, as they first met and fell in love onboard the Titanic. This scene, along with others, portrays their journey of love, loss, and redemption, which ultimately led to their tragic fate. The romantic and emotional connection between Rose and Jack, as well as their shared experiences onboard the Titanic, form the core of their relationship and the movie's themes. 
     \\
    {\normalsize \textcolor{customblue}{\textbf{\lbrack{}End of Assistant 2\rbrack{}}}} \\ \\
    {\normalsize \textcolor{customblue}{\textbf{GPT4-V Output:}}} \\ \\ 
    \textbf{Accuracy:} \\ 
    Scores of the two answers:\\
    Assistant 1: 1 \\ 
    Assistant 2: 7 \\ 
    Reason: Assistant 1 inaccurately describes the scene as featuring a passionate kiss during a storm, which is not depicted in the image. The image shows no kiss or stormy weather; instead, it captures a moment where the characters are standing at the bow of the ship with arms outstretched. This is a significant misinterpretation of the content, leading to a low accuracy score.

    Assistant 2, while not perfectly accurate, describes the scene with less deviation from the image. The mention of them "hanging onto a rope" and it being "a pivotal moment in their story" slightly strays from what is visually depicted, as they are not hanging but rather standing with arms extended and supported. The response is closer to the actual content, thus earning a higher score.
    \\ \\
     \textbf{Detailedness:} \\ 
    Scores of the two answers:\\
    Assistant 1: 3 \\
    Assistant 2: 5 \\ 
    Reason: Assistant 1 provides a narrative with emotional context, describing the intensity and drama of a supposed stormy scene. However, this description does not match the calm, sunset setting in the image and introduces elements like a kiss that aren’t present. While rich in narrative detail, it mismatches the actual visual content significantly.

    Assistant 2 provides more contextual information about the characters and their relationship within the movie, mentioning key themes of love, loss, and redemption. Although it erroneously mentions them "hanging onto a rope" and meeting for the first time in this scene, it still delivers a more aligned description of the emotional and thematic elements present in the actual scene depicted, hence a slightly higher score.
     \\ 
    \end{tabular}
\end{tcolorbox}
\caption{Evaluation example for the image-description evaluator with GPT-4V.}
\label{tab:prompt_example}
\end{minipage}
\end{table*}



\subsection{More case studies}

\begin{figure}[htp] 
    \centering
    \includegraphics[width=\linewidth]{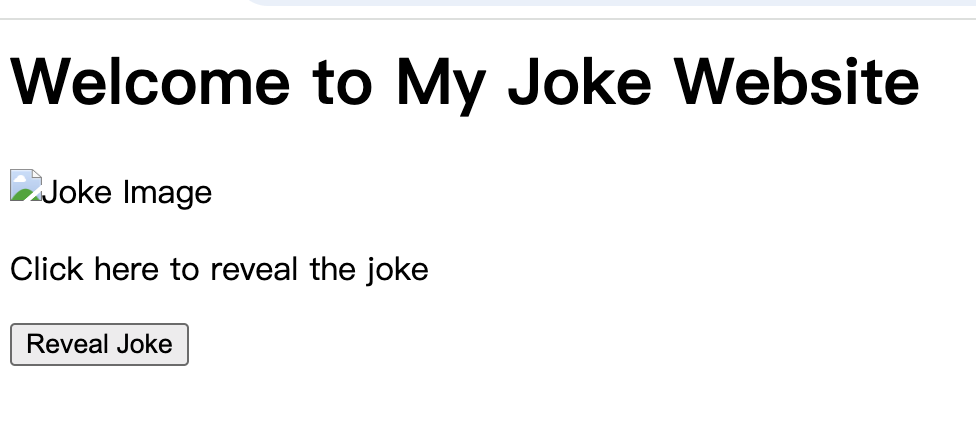}
    \caption{The generated website of \nameshort{}.}
    \label{fig:web}
\end{figure}

\begin{figure*}[h] 
    \centering
    \includegraphics[width=\textwidth]{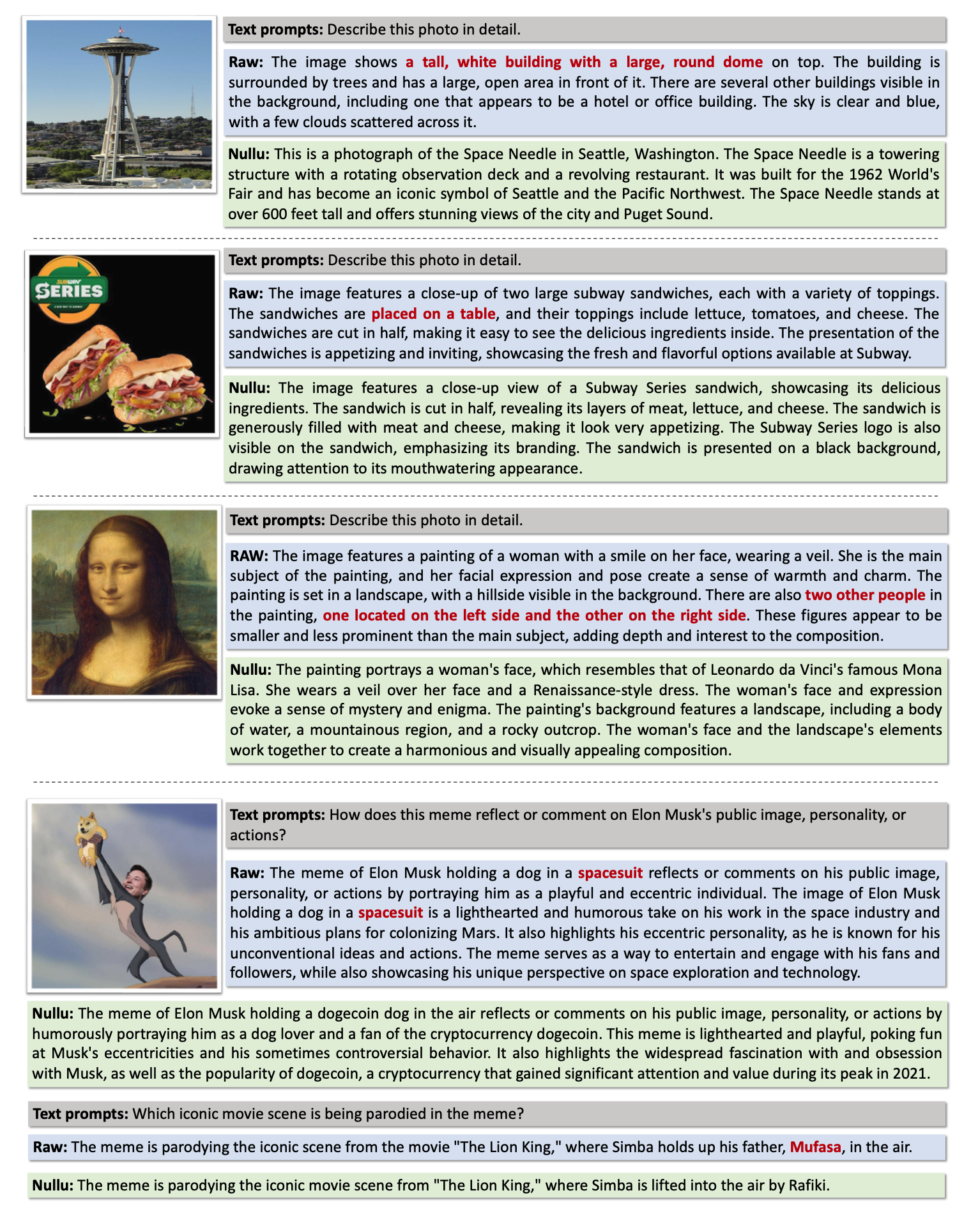}
    \caption{More examples of Nullu on LLaVA-Bench.}
    \label{fig:case1}
\end{figure*}

\begin{figure*}[h] 
    \centering
    \includegraphics[width=\textwidth]{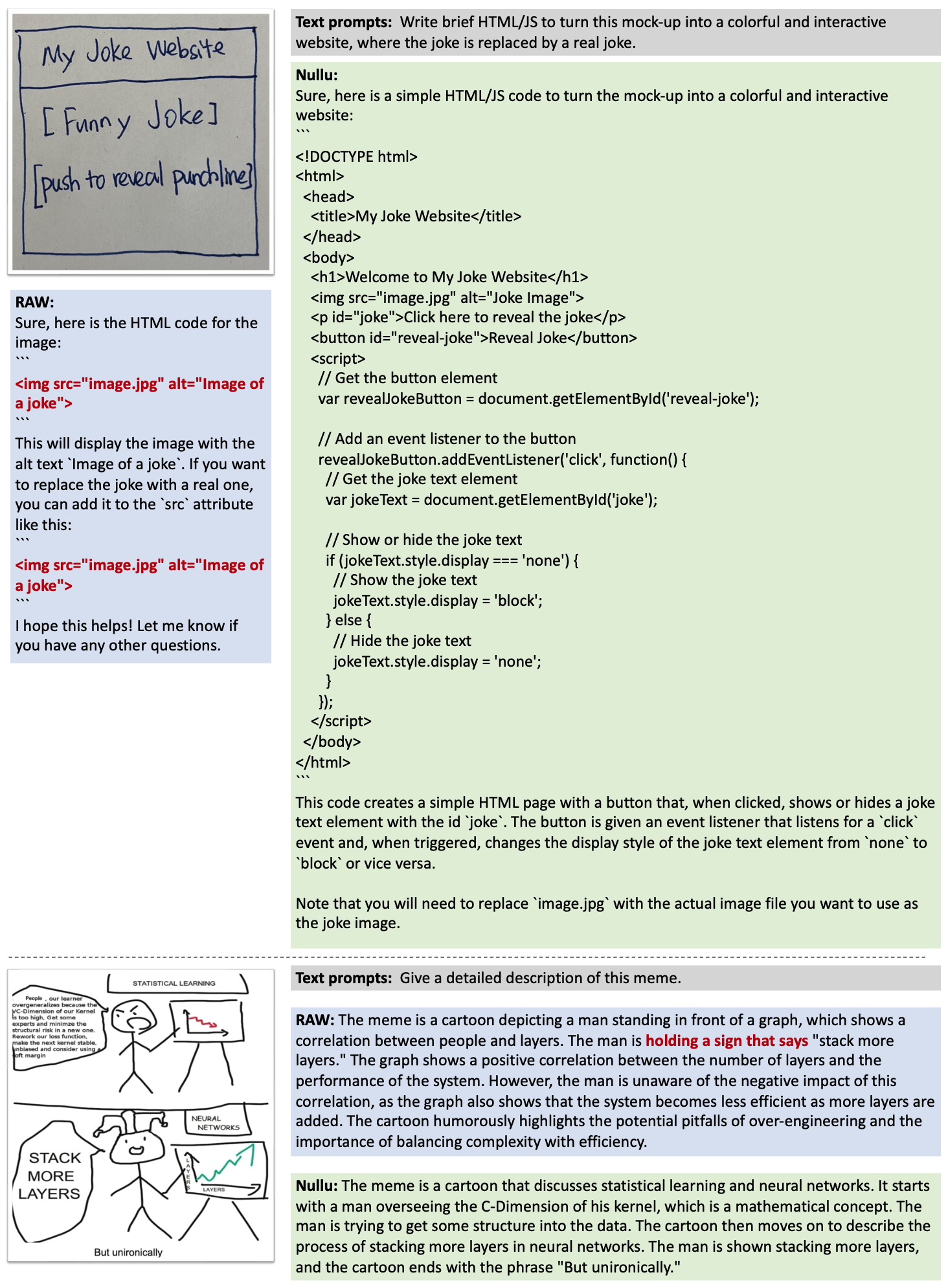}
    \caption{More examples of Nullu on LLaVA-Bench.}
    \label{fig:case2}
\end{figure*}

Additional case studies on the LLaVA-bench are presented in Figure~\ref{fig:case1} and Figure~\ref{fig:case2} to illustrate the effectiveness of our approach. Note that the case in Figure~\ref{fig:case2} provides an example that the proposed \nameshort{} can correctly generate an HTML script that meets the need in the text prompts, which also corresponds to the experimental results on MME, where the edited model is shown that can achieve better performance in Code Reasoning tasks, further demonstrating the effectiveness of our method. The generated HTML website is shown in Figure~\ref{fig:web}. 

\end{document}